\UseRawInputEncoding
\pdfoutput=1

\documentclass[11pt]{article}
\usepackage{natbib}
\usepackage[final]{acl}
\usepackage{tcolorbox}
\tcbuselibrary{skins, breakable, theorems}
\usepackage[frozencache=true,cachedir=minted-cache]{minted} 
\usepackage{times}
\usepackage{multirow}
\usepackage{latexsym}
\usepackage{algorithm,algpseudocode}
\usepackage{booktabs}
\usepackage{amsmath} 
\usepackage{amsfonts} 
\usepackage[T1]{fontenc}
\usepackage{pifont}

\usepackage[utf8]{inputenc}

\usepackage{microtype}
\usepackage{enumitem}
\setlist[itemize]{leftmargin=*,labelsep=0.5em,itemsep=0pt,topsep=0pt}
\usepackage{inconsolata}

\usepackage{graphicx}

\title{PMPO: Probabilistic Metric Prompt Optimization for Small and Large Language Models}

\author{
Chenzhuo Zhao$^1$\thanks{Contributed equally.} \quad 
Ziqian Liu$^2$\footnotemark[1] \quad 
Xinda Wang$^1$ \quad 
Junting Lu$^1$ \\
\textbf{Chaoyi Ruan}$^3$\thanks{Corresponding author.} \\
$^1$Peking University \quad 
$^2$Independent Researcher \quad 
$^3$National University of Singapore \\
\texttt{\{cyzcz, nev\_settle, aidan.lew.37\}@stu.pku.edu.cn} \\
\texttt{liuziqian25@gmail.com} \\
\texttt{ruancy@comp.nus.edu.sg}
}

\begin{document}
\maketitle
~\begin{abstract}
Prompt optimization is a practical and widely applicable alternative to fine tuning for improving large language model performance. Yet many existing methods evaluate candidate prompts by sampling full outputs, often coupled with self critique or human annotated preferences, which limits scalability, especially for smaller models or models that are not instruction tuned. We present PMPO (Probabilistic Metric Prompt Optimization), a unified framework that uses token level cross entropy as a direct, lightweight evaluation signal. PMPO locates low quality prompt segments via a masking based analysis and iteratively rewrites them to propose improved variants. Crucially, during evaluation, PMPO selects among variants by minimizing loss in a single forward pass, eliminating output sampling and human or judge based scoring for selection while still using standard generation only to propose rewrites. This unified, loss based strategy supports both supervised and preference based tasks. Across model sizes and datasets, PMPO outperforms prior prompt optimizers: it achieves the highest average accuracy on BBH, performs strongly on GSM8K and AQuA RAT, and raises AlpacaEval 2.0 win rates by over 19 points. These results demonstrate PMPO’s effectiveness, efficiency, and broad applicability.
\end{abstract}
~\section{Introduction}

\begin{figure}[!ht]
    \centering
    \includegraphics[width=\linewidth]{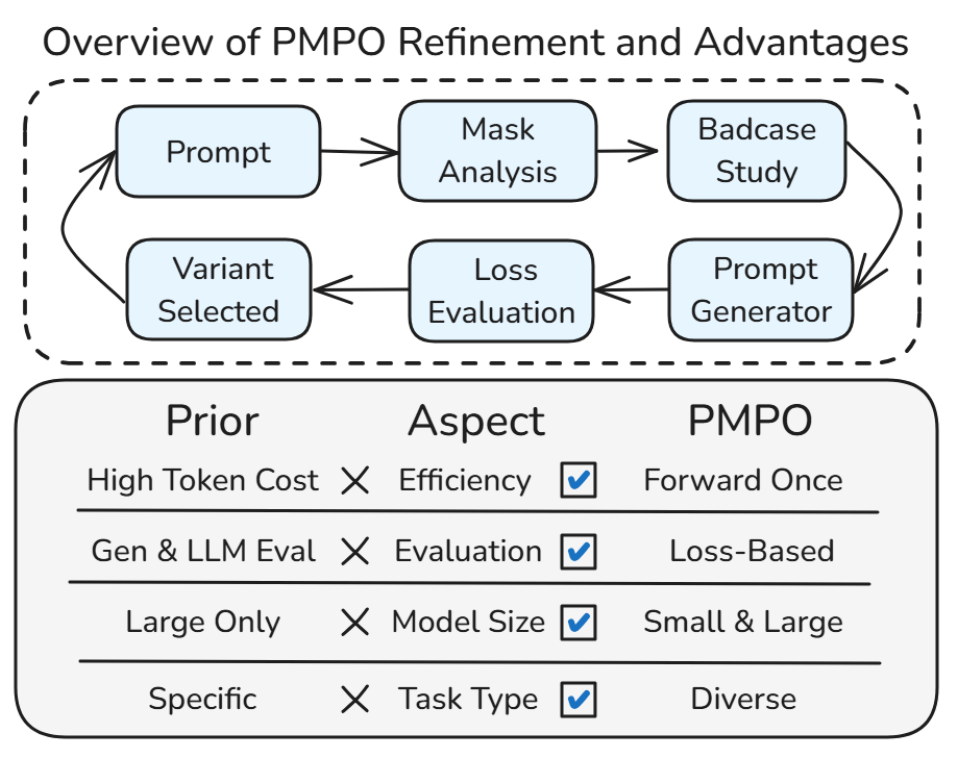}
    \caption{Overview of PMPO. Top: Iterative prompt refinement via loss-based evaluation. Bottom: Comparative strengths over prior methods in terms of evaluation strategy, efficiency, model support, and task generality.}
    \label{fig:pmpo_pipeline}
    \vspace{-0.5em}
\end{figure}

Prompt design has emerged as a critical factor in steering large language models (LLMs) toward reliable performance across diverse tasks. As fine-tuning becomes increasingly costly or restricted , automatic prompt optimization has become a practical alternative to improve model behavior without modifying parameters~\citep{PromptWizard,promptagent}. Recent methods tackle this by iteratively refining prompts through model feedback~\citep{SPO}, human preferences~\citep{bpo}, or reward-based search~\citep{zheng2024pas}. While effective, these approaches often face three main challenges: (1) high cost due to output generation and evaluation loops, (2) reliance on large models capable of introspection or multi-step reasoning, and (3) lack of generality across task types and model sizes.

Some methods, such as PromptWizard~\citep{PromptWizard} and PromptAgent~\citep{promptagent}, rely on model-internal critique and iterative analysis, achieving strong results on complex tasks but requiring significant computation and model cooperation. Others like BPO~\citep{bpo} and PAS~\citep{zheng2024pas} improve efficiency via offline learning of prompt-rewriting models, but require substantial labeled data or preference annotations, and their applicability to new tasks or small models is limited. SPO~\citep{SPO} avoids ground truth by using the model’s own judgment to compare outputs, but this self-evaluation is less reliable for smaller LMs.

A central limitation across these approaches is how prompt quality is evaluated. Most techniques treat prompt evaluation as a generative task, where outputs are generated and then scored externally, making the process computationally expensive and less reliable. External scoring often lacks consistency or sufficient granularity, especially when relying on models to assess complex outputs. Moreover, methods tend to specialize: some focus on supervised tasks with explicit labels, while others target alignment tasks defined by preferences or style. Few offer a unified, efficient mechanism applicable to both.

In this work, we introduce \textbf{PMPO} (Probabilistic Metric Prompt Optimization), a novel prompt optimization framework that directly uses the model’s \textit{cross-entropy loss} as the evaluation signal. PMPO frames prompt optimization as a classic loss minimization problem: given a prompt and input, it computes the likelihood of the desired or preferred output without requiring the model to generate or rank outputs explicitly. This evaluation is efficient, fully automated, and compatible with models of varying scales, making PMPO applicable to both large and small language models in diverse deployment settings. Moreover, PMPO supports both accuracy-based tasks (e.g., classification, QA) and preference-based tasks (e.g., summarization, instruction following) under a unified loss-based evaluation strategy: it minimizes cross-entropy loss for tasks with labeled outputs and maximizes the likelihood of preferred responses over less preferred ones when comparative preference information is available.

Figure~\ref{fig:pmpo_pipeline} presents an overview of the PMPO refinement loop and summarizes its advantages over existing approaches. Compared to prior methods, PMPO offers the following key advantages:
\begin{itemize}
\setlength{\itemsep}{-0.2em} 
    \item \textbf{Loss-based evaluation.} Unlike prior methods, PMPO's evaluation of candidate prompts requires no output sampling or human evaluation, relying only on a single forward pass to compute log-likelihoods, which makes its evaluation process highly efficient.
    \item \textbf{General applicability.} PMPO unifies preference-based and supervised settings by treating prompt optimization as maximizing either output likelihood or reward, depending on the task signal available.
    \item \textbf{Support for small models.} PMPO requires only forward passes and likelihoods, not introspection or reasoning, making it usable for smaller LMs that cannot critique their outputs or support complex prompting.
    \item \textbf{High sample efficiency.} Since loss evaluation is batchable and cheap, PMPO can explore more candidate prompts under fixed budget constraints, enabling extensive search with minimal overhead.
\end{itemize}

~
\begin{figure*}[!ht]
\centering
\includegraphics[width=.9\textwidth]{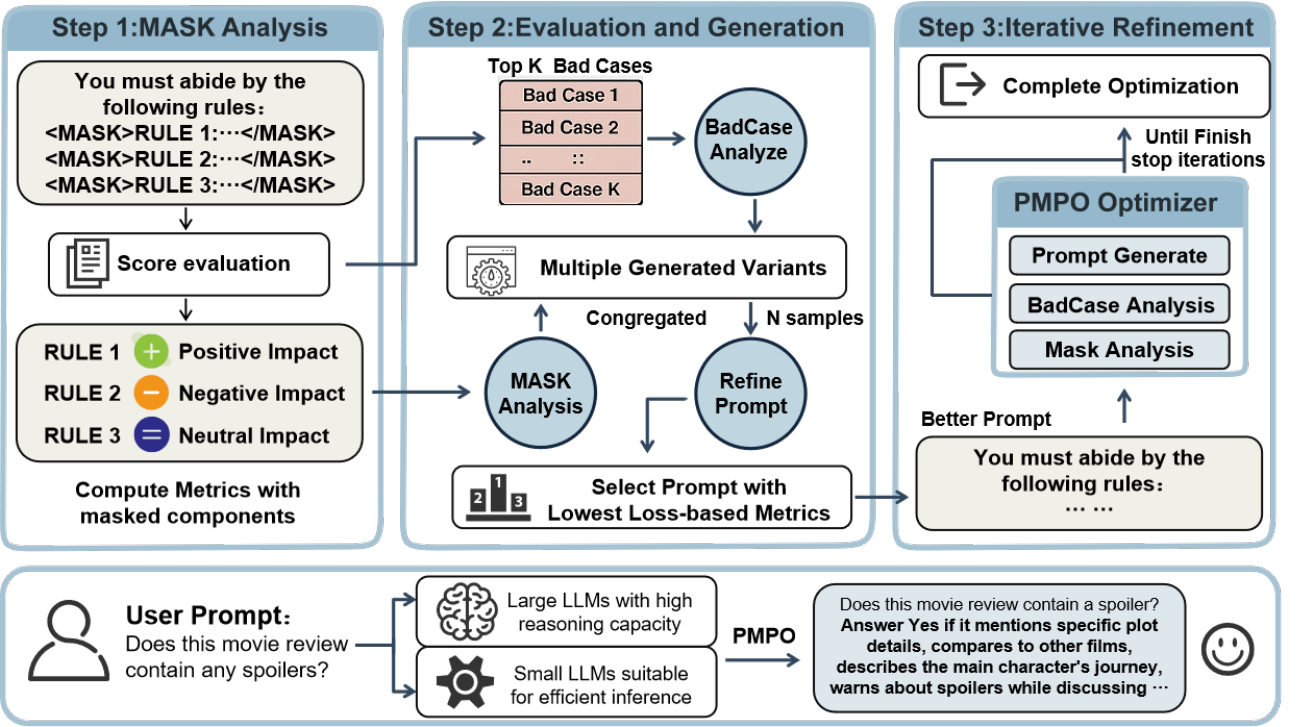}
\caption{Overview of the PMPO framework. Step 1: \textbf{Mask Analysis} estimates the impact of each prompt part by measuring loss changes when masked, identifying positive, negative, or neutral effects. Step 2: \textbf{Evaluation and Generation} selects hard cases and generates refined candidates, scored with the same loss-based metric. The best-performing prompt is retained. Step 3: \textbf{Iterative Refinement} repeats this process until stop.}
\label{fig:framework}
\end{figure*}

\section{Related Works}

\subsection{Prompt Engineering}

Early prompt engineering methods rely on manually crafted instructions and few-shot exemplars to guide language model behavior~\cite{brown2020language,pe1,pe2,pe3}. Chain-of-Thought (CoT) prompting~\cite{cot} enhances this paradigm by encouraging step-by-step reasoning, leading to substantial gains on complex tasks. Other structured prompting methods, such as Step-Back prompting~\cite{zheng2023take}, introduce abstraction or meta-reasoning steps to improve logical inference. Similarly, Rephrase-and-Respond~\cite{rephrase} enhances model comprehension by prompting it to rephrase questions before answering. These techniques improve performance by enriching intermediate reasoning or input interpretation, but they are based on fixed templates and do not perform automated prompt generation or optimization.

In contrast, another line of work explores automatic prompt construction. Methods like APE~\cite{ape} treat prompt design as a search problem, using language models to generate and rank instruction variants based on task performance. While these approaches reduce manual effort, they often rely on large models for generation and evaluation, limiting scalability in low-resource or small-model settings. Overall, this body of work underscores the importance of prompt formulation and motivates the development of more systematic and efficient optimization frameworks.

\subsection{Prompt Optimization}

Recent work on automated prompt optimization can be broadly categorized into generation-based and introspection-based methods~\cite{po_1,po_2,po_ground_2,po_groundtrue}. OPRO~\cite{opro} treats the optimization process as a black-box loop, where an LLM proposes new prompts based on previous attempts and their performance (e.g., accuracy). EvoPrompt~\cite{evoprompt} and PromptBreeder~\cite{promptbreeder} further adopt evolutionary strategies—maintaining populations of prompts that evolve through mutation and selection, with PromptBreeder uniquely co-evolving its mutation strategies. While these approaches often achieve strong performance, they rely on repeated output generation and scoring, which makes them resource-intensive and task-specific. Moreover, their effectiveness depends heavily on large LLMs serving as optimizers, limiting applicability to smaller models or open-ended preference-based tasks.

Another direction leverages the model’s own introspective signals to refine prompts. PromptWizard~\cite{PromptWizard} uses model-internal critiques to iteratively improve prompts and in-context examples. TextGuard~\cite{textguard} adopts a unit-test framework, generating adversarial test cases to probe and correct prompt weaknesses. Though effective in structured tasks with clear correctness criteria, these methods often assume explicit feedback or reference outputs, and require natural-language generation for both diagnosis and revision. Self-supervised approaches like SPO~\cite{SPO} and TextGuard~\cite{textguard} reduce reliance on labels by comparing outputs or generating improvement suggestions, yet still depend on costly output sampling or heuristic reward models.

~
\section{PMPO Framework}
We introduce PMPO, a unified framework designed to iteratively refine prompts using fine-grained cross-entropy evaluation and adaptive rewriting strategies (see Figure~\ref{fig:framework}). PMPO systematically analyzes prompts by applying a model-adaptive masking technique to quantify token-level impacts (Step 1, detailed in Section3.2), and selectively rewrites the lowest-performing segments using language model-generated variants (Step 2, described in Sections3.3 and3.4). Through an iterative cycle of prompt generation, evaluation, and refinement (Step 3, corresponding to the overall loop in Section3.1), PMPO automatically enhances prompt effectiveness without human intervention, adapting flexibly across different model architectures and tasks.

\noindent\textbf{Problem Definition} Given a task $\mathcal{T}$, our framework initializes with a base instruction $\mathcal{P}$ and leverages a problem description paired with a dataset $ \mathcal{D} = {(x_i, y_i, r_i)}_{i=1}^{N} $, where each example consists of an input $x_i$, an output $y_i$, and an associated label or preference score $r_i$ indicating the quality or desirability of $y_i$ for $x_i$. The language model $M$ generates outputs with probabilities $P_M(y \mid x, \mathcal{P})$ conditioned on input $x$ and instruction $\mathcal{P}$. Our objective is to derive an optimal prompt $\mathcal{P}^*$ that maximizes the expected weighted log-probability, that is, $\mathbb{E}_{(x, y, r) \sim \mathcal{D}} \left[ r \cdot \log P_M(y \mid x, \mathcal{P}) \right]$, where $r$ can represent either a binary label or a scalar preference score. Concretely, in accuracy-based tasks (e.g., QA/classification) we set $r_i \equiv 1$, whereas in preference-based tasks we form pairs and assign $r=+1$ to the preferred output and $r=-1$ to the non-preferred one. This unified formulation enables the framework to enhance the model's target-aligned generation capability by amplifying preferences over desired outputs.

\subsection{Iterative Framework for PMPO}

We propose an iterative algorithm that progressively refines prompt quality based on evaluation metrics and targeted modifications. The full procedure is outlined in Algorithm~\ref{alg:ipo}.

\begin{algorithm}[h]
\caption{An Overview of PMPO}
\label{alg:ipo}
\begin{algorithmic}[1]
\Require Dataset $\mathcal{D}$, Initial prompt $\mathcal{P}_0$, Language model $M$, Max iterations $T$, Top-$k$ samples $k$, Variants per sample $n$
\Ensure Optimized prompt $\mathcal{P}^*$
\State $\mathcal{P}^* \leftarrow \mathcal{P}_0$
\For{iteration $\leftarrow 1$ to $T$}
    \State Compute metric $\mathcal{L}(\mathcal{P}^*, M)$ on $\mathcal{D}$
    \State Select top-$k$ samples $\{(x_j, y_j)\}$
    \State Initialize variant set $\mathcal{V} \leftarrow \{\mathcal{P}^*\}$
    \For{each $(x_j, y_j)$}
        \State Analyze failure and token importance using $M$
        \State Generate $n$ variants $\{\mathcal{P}'_{j,1}, ..., \mathcal{P}'_{j,n}\}$
        \State $\mathcal{V} \leftarrow \mathcal{V} \cup \{\mathcal{P}'_{j,1}, ..., \mathcal{P}'_{j,n}\}$
    \EndFor
    \State Evaluate all prompts in $\mathcal{V}$ and select best $\mathcal{P}^{\text{new}}$
    \If{$\mathcal{L}(\mathcal{P}^*, M) > \mathcal{L}(\mathcal{P}^{\text{new}}, M) $}
        \State $\mathcal{P}^* \leftarrow \mathcal{P}^{\text{new}}$
    \EndIf
\EndFor
\State \Return $\mathcal{P}^*$
\end{algorithmic}
\end{algorithm}
In each iteration, the PMPO framework begins with a detailed, mask-guided analysis using the model $M$ to identify specific segments within the current prompt that affect performance. It then computes the cross-entropy or preference-based loss across the entire dataset and selects the most challenging examples, defined as those with the highest losses, as priority targets for improvement. Based on the insights from the mask-guided analysis, the framework generates multiple refined prompt variants, each designed to address the detected weaknesses. These candidates are re-evaluated using batch-level loss performance, and the variant demonstrating the greatest improvement is retained for the next iteration. This refinement process continues until the maximum number of iterations is reached.
\subsection{Mask-Guided Importance Evaluation}
To identify influential components within a instruction $\mathcal{P}$, we first decompose it into a set of semantic units ${s_1, s_2, ..., s_m}$. Instead of relying on predefined rules, we leverage the language model's own reasoning capability to perform this segmentation. This process is driven by a carefully designed meta-prompt, which instructs the model to analyze the instruction and segment it into up to 5 independent and removable semantic units, such as methods, rules, or examples, whose removal would not disrupt the overall coherence of the prompt. The model then wraps these selected segments with \texttt{<mask>...</mask>} tags. A detailed description of our masking formulation and the full meta-prompt template can be found in \textbf{Appendix A.1}.

These identified semantic units are then used to quantify each component's importance. For each unit $s_j$, we create a perturbed prompt variant by masking it:
\begin{equation}
\mathcal{P}^{-j} = \{s_1, s_2, ..., s_{j-1}, \texttt{<MASK>}, s_{j+1}, ..., s_m\}
\end{equation}

We then compute the change in batch-level cross-entropy loss when $s_j$ is masked:
\begin{equation}
\Delta \mathcal{L}_j = \mathcal{L}_{\text{batch}}(\mathcal{P}^{-j}, M) - \mathcal{L}_{\text{batch}}(\mathcal{P}, M)
\end{equation}

Here, $\mathcal{L}_{\text{batch}}(\cdot, M)$ denotes the average cross-entropy loss over the dataset under model $M$. A positive $\Delta \mathcal{L}_j$ indicates that $s_j$ contributes positively to task performance (its removal increases loss), whereas a negative value suggests a detrimental or redundant effect. Values of $\Delta \mathcal{L}_j$ close to zero imply negligible impact on model behavior.

\subsection{Prompt Evaluation via Loss-based Metrics}

To quantitatively assess the effectiveness of prompts, we utilize loss-based metrics derived from the model's internal probability estimates.

Given a model $M$, a instruction $\mathcal{P}$, input $x$, and expected output $y$, we define the token-level cross-entropy loss as:
\begin{equation}
\mathcal{L}_{CE}(x, y, \mathcal{P}, M) = -\sum_{i=1}^{|y|} \log P_M(y_i \mid y_{<i}, x, \mathcal{P})
\end{equation}
where $y_i$ is the $i$-th token of the output sequence and $y_{<i}$ denotes all preceding tokens. The batch-level loss over dataset $\mathcal{D}$ is computed as:
\begin{equation}
\mathcal{L}_{\text{batch}}(\mathcal{P}, M) = \frac{1}{n} \sum_{i=1}^n \mathcal{L}_{CE}(x_i, y_i, \mathcal{P}, M)
\end{equation}

Unlike binary accuracy metrics, $\mathcal{L}_{\text{batch}}$ captures token-wise generation probabilities, provides a continuous evaluation space, and reflects variations in model confidence induced by different prompts.

In scenarios where candidate outputs are associated with preference signals, we additionally incorporate a pairwise preference loss inspired by preference optimization techniques. Given a preferred output $y^+$ and a less preferred alternative $y^-$, we define:
\begin{equation}
\begin{split}
\mathcal{L}_{\text{pref}}(x, y^+, y^-, \mathcal{P}, M) = & \\
-\log \sigma\Big(\beta \cdot \big(s_M(x, y^+, \mathcal{P})& - s_M(x, y^-, \mathcal{P})\big)\Big)
\end{split}
\end{equation}

where $s_M(x, y, \mathcal{P}) = \log P_M(y \mid x, \mathcal{P})$ is the model-assigned log-probability, $\sigma$ is the sigmoid function, and $\beta$ is a scaling factor.

\subsection{Prompt Variant Generation}
To generate new prompt candidates, PMPO employs a model-in-the-loop rewriting mechanism that leverages the language model itself to revise instructions. Rather than using predefined rules or templates, a rewriting prompt is constructed for each selected hard example $(x, y)$ identified by high preference-based cross-entropy loss under the current instruction $\mathcal{P}$. This rewriting prompt includes five key elements: the task description $\mathcal{T}$ to maintain generality, the current instruction $\mathcal{P}$ as the base for revision, the hard example $(x, y)$ to expose weaknesses, editing instructions focused on improving clarity, specificity, and structural quality, and a token-level mask analysis that highlights segments in $\mathcal{P}$ contributing most to the loss.

The mask-guided analysis functions as a soft signal rather than a hard pruning mechanism. It does not constrain the optimization to specific regions; it merely suggests areas where targeted edits may be beneficial. During candidate generation, the model is encouraged to explore multiple rewriting variants that are not strictly limited to the masked regions, ensuring comprehensive exploration. This approach also confers robustness: even if the initial LLM-based segmentation is imperfect, the framework is iterative and combines variant selection with evaluation to enable continuous improvement and refinement across multiple cycles.

The model is guided to first diagnose flaws in $\mathcal{P}$ and then apply a multi-step refinement strategy. These steps include rephrasing rigid wording, refining task constraints, removing redundancy, simplifying overly complex instructions, improving logical flow, expanding underspecified parts, merging overlapping rules, and enhancing overall language quality. These edits are applied in an adaptive and integrated manner based on the specific issues observed, preventing overfitting to individual cases and encouraging broadly effective improvements. Each hard example $(x_j, y_j)$ yields a set of revised prompts ${\mathcal{P}'{j,1}, \ldots, \mathcal{P}'{j,n}}$ generated via temperature-controlled top-$p$ sampling. These variants are pooled and evaluated using batch-level cross-entropy, and the best-performing candidate, measured by loss or accuracy, is selected as the updated prompt for the next iteration.


\subsection{Efficiency of PMPO}
A core advantage of PMPO is its computational and architectural efficiency across both evaluation and candidate prompt generation. For each prompt variant under consideration, PMPO directly leverages token level likelihoods, computing cross entropy over the target outputs in a single forward pass per example and thereby bypassing costly autoregressive decoding. While the overall optimization framework is iterative, cycling through prompt generation, selection, and refinement, this highly efficient per candidate scoring enables rapid evaluation of many variants under a fixed compute budget without relying on external classifiers, preference models, or second pass LLM judges.
This enables batched scoring of many prompt candidates even under limited compute. Prompt rewriting is likewise streamlined: for each high-loss example, the model performs a one-shot analysis and proposes revised prompt variants without iterative reasoning, multi-step prompting, or introspective feedback. As summarized in Table~\ref{tab:efficiency_comparison}, this combination of token-level evaluation and multi-candidate scoring yields a favorable efficiency profile relative to generation-heavy methods such as OPRO or PromptWizard, whose multi-step refinement loops are computationally intensive and can underperform on smaller models with limited reasoning capacity~\cite{zhang2024revisitingoprolimitationssmallscale, mayilvaghanan2025propel}.

\begin{table}[!ht]
\centering
\small

\resizebox{\linewidth}{!}{
\begin{tabular}{l|c|c}
\hline
\textbf{Method} & \textbf{Evaluation Level} & \textbf{Candidate Count} \\
\hline
SPO & Sequence-Level & Single \\
PromptWizard & Sequence-Level  & Multiple \\
EvoPrompt & Sequence-Level & Multiple \\
OPRO & Sequence-Level & Single \\
\textbf{PMPO} & Token-Level & Multiple \\
\hline
\end{tabular}
}
\caption{Comparison of Efficiency.}
\label{tab:efficiency_comparison}
\end{table}

~\begin{table*}[!ht]
\centering
\small
\caption{
Average test accuracy in the 1-shot setting across multiple tasks for different prompting methods, evaluated on Qwen2.5-14B-Instruct. Bold values indicate the best-performing method for each task.
}
\setlength{\tabcolsep}{3.5pt}
\renewcommand{\arraystretch}{1.2}
\begin{tabular}{lcccc|ccc|c}
\hline
\textbf{Task Name} & \textbf{AO} & \textbf{CoT} & \textbf{RaR} & \textbf{StepBack} & \textbf{OPRO} & \textbf{EvoPrompt} & \textbf{PromptWizard} & \textbf{Ours} \\
\hline
boolean\_expressions & 0.756 & 0.920 & 0.952 & 0.936 & 0.972 & 0.952 & \underline{0.976} & \textbf{0.984} \\
causal\_judgement & \underline{0.674} & 0.631 & \textbf{0.695} & 0.658 & 0.636 & 0.647 & 0.599 & \textbf{0.695} \\
date\_understanding & 0.684 & 0.740 & 0.708 & 0.752 & \textbf{0.800} & 0.772 & 0.636 & \underline{0.784} \\
disambiguation\_qa & 0.656 & 0.776 & 0.716 & 0.640 & \underline{0.848} & 0.760 & \textbf{0.892} & 0.736 \\
dyck\_languages & 0.096 & 0.240 & 0.236 & 0.228 & \textbf{0.392} & \underline{0.308} & 0.220 & 0.256 \\
formal\_fallacies & 0.704 & 0.800 & 0.784 & 0.808 & \textbf{0.856} & 0.792 & \underline{0.816} & \underline{0.816} \\
geometric\_shapes & 0.440 & 0.616 & 0.576 & \textbf{0.684} & 0.580 & 0.620 & 0.508 & \underline{0.676} \\
hyperbaton & 0.632 & 0.704 & 0.768 & \underline{0.848} & 0.740 & 0.756 & 0.752 & \textbf{0.896} \\
logical\_deduction & 0.692 & 0\underline{.856} & 0.844 & 0.847 & 0.767 & \textbf{0.864} & 0.845 & \textbf{0.864} \\
movie\_recommendation & 0.564 & 0.636 & 0.624 & 0.640 & 0.636 & 0.596 & \underline{0.676} & \textbf{0.684} \\
multistep\_arithmetic\_two & 0.052 & 0.968 & 0.972 & 0.956 & \textbf{0.988} & 0.948 & \underline{0.976} & \textbf{0.988} \\
navigate & 0.660 & 0.908 & 0.856 & 0.924 & 0.896 & \underline{0.944} & 0.848 & \textbf{0.960} \\
object\_counting & 0.508 & 0.812 & 0.772 & 0.756 & 0.688 & \underline{0.876} & 0.832 & \textbf{0.884} \\
penguins\_in\_a\_table & 0.753 & 0.945 & 0.932 & \underline{0.952} & 0.932 & \textbf{0.959} & 0.726 & \underline{0.952} \\
reasoning\_about\_colored\_objects & 0.708 & \underline{0.892} & 0.768 & 0.880 & 0.876 & 0.872 & \textbf{0.912} & 0.888 \\
ruin\_names & 0.632 & 0.660 & 0.556 & 0.716 & \underline{0.788} & 0.680 & 0.692 & \textbf{0.840} \\
salient\_translation\_error\_detection & 0.600 & 0.572 & 0.604 & \textbf{0.644} & \underline{0.624} & 0.604 & 0.504 & 0.600 \\
snarks & 0.831 & 0.809 & 0.837 & \underline{0.848} & 0.843 & \textbf{0.882} & 0.787 & 0.826 \\
sports\_understanding & 0.752 & 0.660 & 0.680 & 0.804 & \underline{0.828} & 0.812 & 0.544 & \textbf{0.836} \\
temporal\_sequences & 0.832 & 0.908 & 0.864 & 0.900 & \textbf{0.964} & 0.916 & 0.900 & \underline{0.944} \\
tracking\_shuffled\_objects & 0.599 & \textbf{0.900} & 0.852 & 0.847 & 0.860 & 0.871 & 0.839 &\underline{0.880} \\
web\_of\_lies & 0.536 & 0.900 &\underline{0.972} & 0.920 & 0.820 & 0.900 & 0.716 & \textbf{0.976} \\
word\_sorting & 0.276 & 0.444 & \textbf{0.624} & 0.600 & 0.388 & \underline{0.608} & 0.544 & 0.580 \\
\hline
\textbf{Best performing tasks} & 0 & 1 & 2 & 2 & \underline{5} & 3 & 2 & \textbf{11} \\
\textbf{Average Accuracy} & 0.593 & 0.752 & 0.747 & 0.773 & 0.770 & \underline{0.780} & 0.728 & \textbf{0.806} \\
\hline
\end{tabular}
\label{tab:ours_1shot_comparison}
\end{table*}

\begin{table}[!ht]
\centering
\small
\setlength{\tabcolsep}{6pt}
\renewcommand{\arraystretch}{1.15}
\begin{tabular}{l|c|c}
\hline
\textbf{Method} & \textbf{GSM8K} & \textbf{AQUA-RAT} \\
\hline
AO & 0.871 & 0.760 \\
APE & \underline{0.939} & 0.827 \\
COT & 0.907 & \underline{0.843} \\
RaR & 0.932 & \underline{0.843} \\
Step-back & 0.925 & 0.811 \\
\hline
OPRO & 0.936 & 0.819 \\
PromptBreeder & 0.917 & 0.831 \\
PromptWizard & 0.882 & 0.799 \\
Textguard & 0.939 & 0.807 \\
\hline
\textbf{Ours} & \textbf{0.940} & \textbf{0.846} \\
\hline
\end{tabular}
\caption{Accuracy on math reasoning datasets GSM8K and AQUA-RAT using different prompt optimization methods(0-Shot), use Qwen2.5-14B as their optimization models.}
\label{tab:math_results}
\vspace{-0.5em}
\end{table}

~\section{Experiment}
\subsection{Experiment Settings}

\noindent\textbf{Dataset.} We evaluate PMPO on a diverse set of benchmarks covering mathematical reasoning, logical inference, and open-ended instruction following. For math problem solving, we use \textbf{GSM8K} and \textbf{AQUA-RAT}, which require models to perform multi-step numerical reasoning. To assess logical reasoning capabilities, we adopt the \textbf{BBH} benchmark, designed to challenge models with complex inference tasks under minimal guidance. For evaluating general instruction-following and open-ended task performance, we use \textbf{AlpacaEval 2.0}, which includes a broad range of user instructions and uses GPT-4 Turbo as an automatic evaluator to compare model responses against reference answers.

\noindent\textbf{Baseline.} We compare PMPO against two categories of prompting methods across benchmark datasets. The first category includes conventional manually designed prompting strategies: CoT, RaR, and StepBack, which enhance model reasoning or answer formulation through structural or logical heuristics. The second category comprises recent automated prompt optimization approaches: OPRO, EvoPrompt, PromptWizard, TextGuard, and PromptBreeder, which leverage language models to search, mutate, or iteratively improve prompts without human intervention.

\noindent\textbf{Implementation Details.}  We conduct experiments using a mix of open-source and proprietary language models, including Qwen2.5 (0.5B, 14B, 32B), LLaMA3.1 (8B), DeepSeek-R1-Distill-Qwen (1.5B). For each dataset, we randomly select 20\% of the examples for training (capped at 50) and evaluate on the remaining set. The preference scaling factor $\beta$ is fixed at 1 across all experiments.
In each optimization round, we choose the top-$k = 3$ most challenging samples and generate 4 prompt variants per sample. Optimization runs for up to 20 iterations. All experiments are conducted on a single NVIDIA H800 GPU, with each full optimization taking approximately 20 minutes.

\subsection{Experimental Results and Analysis}

\noindent\textbf{Reasoning Tasks.} As shown in Table~\ref{tab:ours_1shot_comparison}, our method consistently outperforms existing prompting approaches across a wide range of reasoning and understanding tasks, evaluated on Qwen2.5-14B-Instruct under a 1 shot setting for fair comparison. PMPO achieves the highest average accuracy of 80.6\%, surpassing strong baselines such as EvoPrompt (78.0\%), OPRO (77.1\%), and PromptWizard (72.8\%), and ranks first on 11 out of 23 tasks, significantly more than any other method. Unlike generation heavy strategies like OPRO and EvoPrompt, which use black box search or population based mutation, PMPO’s single pass, loss based evaluation enables more efficient optimization and stronger performance, especially on tasks that require multi step or spatial reasoning. While PromptWizard is competitive on some logic tasks, its self critique based rewrites often create verbose, rigid prompts that limit generalization. In contrast, PMPO produces lightweight and structurally adaptive prompts that more effectively align with model behavior, yielding robust performance across both symbolic and naturalistic reasoning.


\noindent\textbf{Math Tasks.} As shown in Table~\ref{tab:math_results}, our method achieves the best performance on two widely used math reasoning benchmarks, GSM8K and AQUA-RAT, with accuracies of 94.0\% and 84.6\% respectively. It outperforms all baselines, including APE, CoT (90.7\% on GSM8K, 84.3\% on AQUA-RAT), and PromptBreeder. Notably, both datasets provide complete solution steps rather than just final answers, which we incorporate as positive targets during training. This enables PMPO to iteratively refine prompts using token-level confidence (cross-entropy loss), effectively guiding the model to generate detailed, accurate multi-step reasoning. In contrast to methods like Chain-of-Thought that rely on manually designed heuristics, PMPO’s approach not only enhances final answer accuracy but also significantly improves the quality and fidelity of intermediate steps. It encourages the model to emulate human-like problem-solving behaviors such as decomposition, variable definition, and numeric justification, thereby demonstrating stronger alignment with complex mathematical reasoning tasks while maintaining generalizability.

\begin{table*}[!ht]
\centering
\small
\setlength{\tabcolsep}{5pt}
\renewcommand{\arraystretch}{1.15}
\begin{tabular}{l|c|c|c|c|c}
\hline
\shortstack{\textbf{Target $\backslash$} \\ \textbf{Prompt Source }}& 
\shortstack{Qwen2.5\\0.5B} & 
\shortstack{DeepSeek\\1.5B} & 
\shortstack{LLaMA\\3.1–8B} & 
\shortstack{Qwen2.5\\14B} & 
\shortstack{Qwen2.5\\32B} \\
\hline
\shortstack{Qwen2.5-0.5B}       & \textbf{0.580} & 0.568 & 0.464 & 0.500 & \underline{0.580} \\
\shortstack{DeepSeek-1.5B}      & 0.612 & \textbf{0.772} & \underline{0.700} & 0.640 & 0.584 \\
\shortstack{LLaMA-3.1–8B}         & 0.708 & 0.792 & 0.800 & \underline{0.852} & \textbf{0.860} \\
\shortstack{Qwen2.5-14B}        & 0.912 & 0.948 & 0.896 & \textbf{0.960} & \underline{0.956} \\
\shortstack{Qwen2.5-32B}        & 0.972 & 0.948 & 0.952 & \underline{0.972} & \textbf{0.980} \\
\hline
\end{tabular}
\caption{Cross-model accuracy on Navigate. Each row corresponds to a \textbf{target model} and each column to the \textbf{prompt source model}. Diagonal entries represent same-model optimization. Models are ordered from smallest to largest.}
\label{tab:cross_model}
\end{table*}

\begin{figure}[ht]
    \centering
    \includegraphics[width=0.90\linewidth]{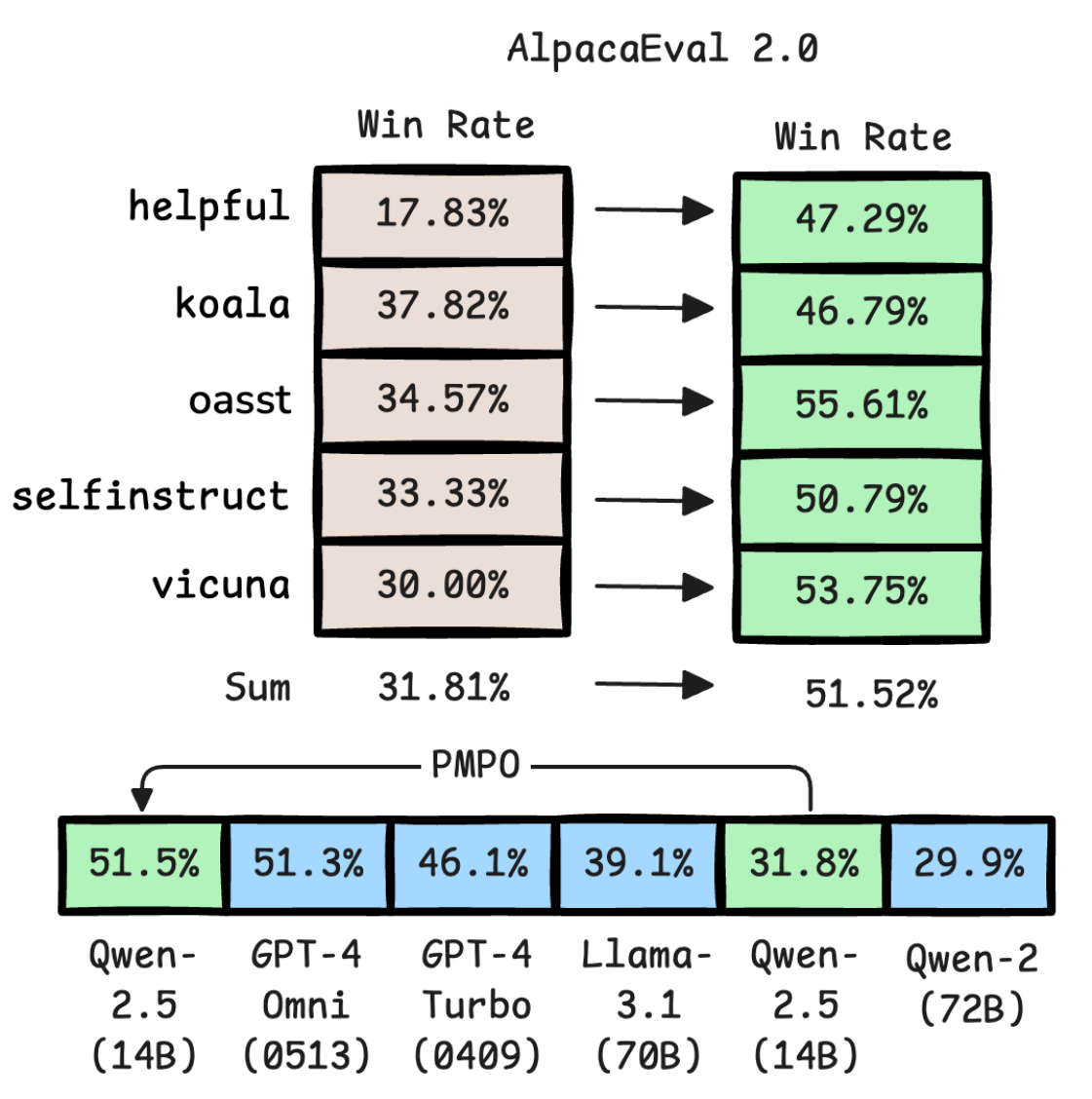}
    \caption{Win rate comparisons on AlpacaEval 2.0. Left: original Qwen2.5-14B win rates across instruction sources. Right: PMPO-optimized results using the same model. Bottom: average win rates across models.}
    \label{fig:alpacaeval}
\end{figure}

\noindent \textbf{Open-ended Dataset.} To evaluate PMPO on open-ended instruction-following tasks, we conduct experiments on the AlpacaEval 2.0 benchmark. Using GPT-4 Turbo as the evaluator, we compare model-generated responses with reference outputs across five instruction categories. As shown in Figure~\ref{fig:alpacaeval}, PMPO-optimized prompts raise the average win rate of Qwen2.5-14B from \textbf{31.81\%} to \textbf{51.52\%}, a substantial improvement. Notably, our method boosts performance across all instruction subsets, including difficult ones like \texttt{helpful} (from 17.83\% to 47.29\%) and \texttt{oasst} (from 34.57\% to 55.61\%).

These results show that PMPO enables mid-sized models like Qwen2.5-14B to produce competitive outputs, surpassing larger models such as LLaMA3.1-70B (39.1\%) and GPT-4 Turbo (46.1\%), and nearly matching GPT-4 Omni (51.3\%). This highlights PMPO’s effectiveness in enhancing instruction alignment without model fine-tuning or explicit preference labels.

\subsection{Open-Source Cross-Model Generalization Analysis }
To assess adaptability across model scales and architectures, we conduct cross model evaluation by applying prompts optimized on one model to others (Table~\ref{tab:cross_model}, BBH:Navigate). Models of various sizes, including small, medium, and large, are evaluated in a zero shot manner using the optimized prompts for inference.

Results reveal a notable instruction-following capacity gap: prompts optimized on large models, though effective on similarly large or medium models, often degrade when transferred to smaller ones, which may struggle with complex or verbose instructions. Furthermore, prompts consistently perform best on the originating model, indicating that prompt effectiveness is closely linked to the internal reasoning and instruction-following mechanisms of each model.

\begin{figure}[ht]
    \centering
    \includegraphics[width=1.0\linewidth]{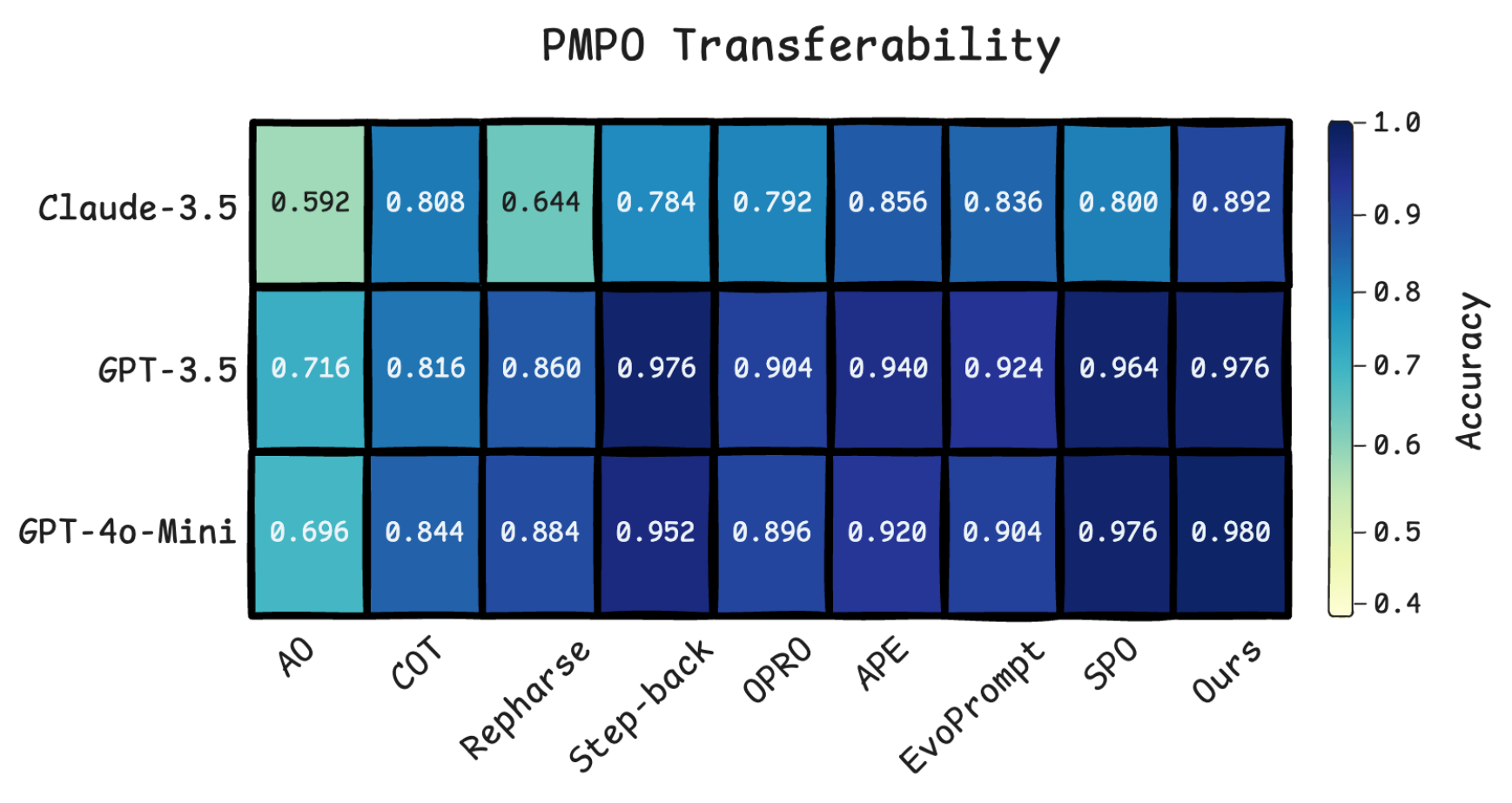}
    \caption{Cross-model performance on BBH:Navigate using prompts optimized on Qwen2.5-32B.}
    \label{fig:cross_model_transfer}
\end{figure}

\subsection{Transferability to Proprietary Models}

While PMPO is fully applicable to open-source models that support token-level likelihood access, applying it to proprietary systems introduces additional constraints. These models typically provide log-probabilities only for generated tokens, without supporting full-sequence evaluation through a single forward pass. As a result, it is not straightforward to compute cross-entropy loss over an entire dataset. One potential workaround is to construct step-wise input sequences by incrementally appending target tokens and querying the model at each step to obtain token-level probabilities. However, this procedure can lead to substantial token consumption and high latency. Therefore, we do not recommend applying the full PMPO optimization process directly on these models in practice.

Despite the limited access to loss signals on proprietary systems, prompts optimized via PMPO on open-source models still demonstrate strong cross-model transferability. Although optimized on Qwen2.5-32B, these prompts consistently improve performance on the BBH:Navigate benchmark when applied to GPT-3.5 Turbo 0613, Claude 3.5 Haiku 20241022, and GPT-4o (Figure~\ref{fig:cross_model_transfer}).

\subsection{Reasoning Process Fidelity}
To quantitatively validate our claim that PMPO improves the fidelity of intermediate reasoning steps, we conducted an additional analysis on the GSM8K benchmark using a specialized Process Reward Model (PRM), \textbf{Qwen2.5-Math-PRM-7B}. This model is specifically trained to assess the quality of individual steps in mathematical solutions. For each prompting method, we parsed the generated solutions into individual reasoning steps using the newline character as a delimiter. Each step was then independently scored by the PRM, and we computed the average score across all steps to determine the final process reward for each sample.

As shown in Table~\ref{tab:prm_scores}, PMPO achieves the highest average process reward score (0.9950), indicating that the reasoning steps it generates are of the highest quality among all compared methods. This result provides direct, quantitative evidence that PMPO not only excels in final answer accuracy but also generates more reliable and logically sound intermediate steps. This validates our claim that PMPO's loss-guided refinement encourages the model to emulate more structured and correct human-like problem-solving behaviors.

\begin{table}[ht]
\centering
\small
\begin{tabular}{l|c}
\hline
\textbf{Prompting Method} & \textbf{Process Reward Score} \\
\hline
AO & 0.5510 \\
PromptWizard & 0.8867 \\
CoT & 0.9637 \\
PromptBreeder & 0.9776 \\
RaR & 0.9910 \\
OPRO & 0.9926 \\
APE & 0.9930 \\
Step-back & 0.9932 \\
TextGrad & 0.9940 \\
\textbf{PMPO (Ours)} & \textbf{0.9950} \\
\hline
\end{tabular}
\caption{Process reward scores on GSM8K evaluated by Qwen2.5-Math-PRM-7B. Higher scores indicate better intermediate reasoning quality. PMPO achieves the highest score.}
\label{tab:prm_scores}
\end{table}

\subsection{Ablation Study}
To assess the contribution of individual components within PMPO, we conduct a cumulative ablation study on the BBH benchmark using Qwen2.5-14B for both optimization and evaluation. Holding the iterative refinement mechanism fixed, we successively ablate Token Importance Masking (TIM), Bad Case Analysis (BCA), and Preference Loss (PrefLoss). TIM localizes low-performing prompt spans via token-level loss attribution, enabling edits to concentrate where the model struggles; removing it drops accuracy from 80.63\% to 79.05\%. Eliminating BCA, which prioritizes high-loss examples for targeted refinement, yields a further decline to 77.96\%. Finally, omitting PrefLoss, our preference-learning objective used to rank and select candidate rewrites, reduces accuracy to 76.74\%.

The strictly monotonic degradation observed in Table~\ref{tab:compact_ablation} indicates that each module makes a non-redundant, complementary contribution: TIM identifies where to edit, BCA focuses which cases to fix first, and PrefLoss guides how to prefer improved rewrites. Together, these mechanisms deliver a synergistic gain beyond the core iterative loop, underscoring the importance of fine-grained evaluation and targeted rewriting for maximizing performance.
\begin{table}[ht]
\centering
\small
\setlength{\tabcolsep}{10pt}
\resizebox{\linewidth}{!}{%
\begin{tabular}{l|c|c|c|c}
\hline
\textbf{Setting} & \textbf{TIM}  & \textbf{BCA}& \textbf{PrefLoss} & \textbf{Acc. (\%)} \\
\hline
Full & \ding{51} & \ding{51} & \ding{51} & \textbf{80.63} \\
w/o TIM & \ding{55} & \ding{51} & \ding{51} & 79.05 \\
w/o TIM,BCA & \ding{55} & \ding{55} & \ding{51} & 77.96 \\
w/o TIM,BCA,PrefLoss & \ding{55} & \ding{55} & \ding{55} & 76.74 \\
\hline
\end{tabular}
}
\caption{Ablation results on BBH dataset using Qwen2.5-14B. Removing individual modules leads to progressive performance degradation.}
\label{tab:compact_ablation}
\end{table}

~

\section{Conclusion}
We present PMPO, a unified and efficient framework for prompt optimization that relies on loss-based evaluation and iterative rewriting. Instead of relying on output generation or human feedback, PMPO uses token-level likelihoods to identify and refine underperforming prompt segments, enabling scalable and efficient optimization. Experimental results on reasoning, mathematical, and instruction-following benchmarks show that PMPO consistently outperforms existing methods in both accuracy and efficiency. Its lightweight design, compatibility with smaller models, and minimal need for manual supervision make it well-suited for both academic research and practical deployment.



\section*{Limitations}
Despite the promising results, our study has several imitations. 

While PMPO demonstrates strong efficiency and effectiveness in optimizing prompts across a range of open-source models, its application to proprietary, closed-source language models remains limited. Most commercial APIs (e.g., OpenAI, Anthropic) do not expose full log-likelihoods, which restricts the direct use of PMPO’s loss-based evaluation due to privacy constraints and concerns around model behavior leakage. Although approximate likelihoods can be estimated via autoregressive token-by-token querying, this significantly increases latency and token usage, making it impractical for large-scale optimization. Nonetheless, some API-based frameworks, such as vLLM, do provide access to token-level log-probabilities (e.g., via \texttt{prompt\_logprobs}), allowing PMPO to be applied in those settings. We view this as a positive direction, and hope that more commercial providers will consider offering similar transparency to facilitate research on prompt optimization and model alignment.

Additionally, in extremely low-resource scenarios (e.g., using only one training example), PMPO may exhibit reduced robustness. Since the optimization directly minimizes model loss on a limited number of instances, it is prone to overfitting in such cases. If no additional data is introduced, the resulting prompt variants may align too closely with the few observed examples, leading to reduced generalization. While this setup is inherently challenging for most learning algorithms, it highlights a fundamental limitation of data-scarce prompt optimization.
\bibliography{custom}

\clearpage
\appendix
\onecolumn
\section{Appendix}
\label{sec:appendix}

~
\renewcommand{\thetable}{A\arabic{table}}
\renewcommand{\thefigure}{A\arabic{figure}}
\subsection{Detailed Prompts of ours}
\label{appendix:prompt}

In this section, we present the meta prompts used in ours for prompt optimization, evaluation, and segment-level masking. These prompts enable the framework to analyze, rewrite, and assess instructions with minimal human intervention. We include variants adapted for large and small models, as well as the masking strategy used for loss attribution.

\vspace{0.5em}
\noindent\textbf{Prompt Optimization (Large Model).}  
We first show the optimization prompt used with capable instruction-following LMs:

\begin{tcolorbox}[title={\textbf{\small Optimization Prompt for Large Models}}, boxrule=2pt, arc=0mm, breakable]
\begin{minted}[fontsize=\scriptsize, breaklines, frame=lines, framesep=2mm, tabsize=4, style=vs, autogobble]{python}
You are an expert prompt engineer tasked with dynamically improving prompts to generate more effective, diverse solutions. When analyzing a prompt, first diagnose its core weaknesses, then apply multiple strategic modifications as needed.

Given the current prompt and task description, your objective is to produce a significantly improved version that will better solve the intended task.

Your optimization must center on the task description: {task_description}

CRITICAL WARNING - MAINTAIN TASK SCOPE:
    1. The task_description defines the FULL SCOPE of what your prompt must address
    2. Examples are provided ONLY to understand the FORMAT, not to narrow the task
    3. Your prompt MUST maintain the original breadth of the task_description
    4. NEVER specialize the prompt to only handle specific examples you've seen

    Example Input: {user_input}
    Expected Answer: {true_answer}

These examples are provided ONLY for pattern analysis. Do NOT directly incorporate these exact examples into your prompt or design your prompt specifically for these examples. Instead:
- Extract the underlying patterns and reasoning these examples demonstrate
- Understand the general skills or knowledge being tested
- Focus your prompt improvements on the task_description and solving the general problem

First, analyze the prompt for:
- Gaps in instruction clarity or specificity
- Unnecessary constraints limiting creative problem-solving
- Missing guidance that would help solve the general task type
- Overly rigid structure that hinders diverse approaches
- Places where more natural, professional language would improve understanding
- Redundancies or contradictions causing confusion

Then, apply a strategic combination of these techniques (using multiple approaches rather than just one):
(1) ENHANCE STRUCTURE, (2) ADD RULES OR PRINCIPLES, (3) REMOVE UNNECESSARY ELEMENTS,
(4) REPHRASE FOR CLARITY, (5) SIMPLIFY COMPLEXITY, (6) EXPAND WITH DETAILS,
(7) CONSOLIDATE SIMILAR RULES, (8) PROFESSIONAL REFRAMING, (9) DIVERSIFY APPROACH

IMPORTANT: Your response must focus on creating a prompt that will produce substantively better results on the general task, not just on the specific examples provided.

Additionally, here is an analysis of the current prompt, segmented by mask (for reference in your optimization):  
{mask}

Current prompt:  
{current_prompt}

For the final prompt, please wrap it with <prompt></prompt>.
\end{minted}
\end{tcolorbox}

\vspace{0.5em}
\noindent\textbf{Prompt Optimization (Small Model).}  
To accommodate smaller LMs with limited instruction-following capacity, we use a simplified variant:

\begin{tcolorbox}[title={\textbf{\small Optimization Prompt for Small Models}}, boxrule=2pt, arc=0mm, breakable]
\begin{minted}[fontsize=\scriptsize, breaklines, frame=lines, framesep=2mm, tabsize=4, style=vs, autogobble]{python}
Task_description: {task_description}
Example Input: {user_input}
Expected Answer: {true_answer}
Mask Analysis: {mask}

Current Prompt: {current_prompt}

Based on the following examples where the current cross-entropy is relatively high, please analyze the reasons and modify the prompt to improve performance. Rather than directly quoting the examples, focus on deeply analyzing the underlying patterns and issues that contribute to high cross-entropy. Prioritize identifying the root causes of performance problems and make labeled prompt modifications that address these specific issues. Concentrate on what changes will most effectively improve task outcomes rather than structural coherence or theoretical correctness.

Your response must focus on creating a prompt that will produce substantively better results on the general task, not just on the specific examples provided.

Please only wrap the optimized final prompt with <prompt></prompt> tags.
\end{minted}
\end{tcolorbox}

\vspace{0.5em}
\noindent\textbf{Prompt Evaluation.}  
To evaluate candidate outputs, we use a lightweight semantic comparison prompt that tolerates format variation:

\begin{tcolorbox}[title={\textbf{\small Evaluation Prompt}}, boxrule=2pt, arc=0mm, breakable]
\begin{minted}[fontsize=\scriptsize, breaklines, frame=lines, framesep=2mm, tabsize=4, style=vs, autogobble]{python}
You are an expert evaluator determining if an answer matches the ground truth. Consider equivalent formats like 'A', '(A)', 'A.', etc. as correct. Focus on the meaning rather than exact string matching. For questions where the answer format is important, verify that the model answers in the correct format. For example, in Dyck language problems, if the question asks what follows '([{}'  and the ground truth is '])', but the model answers with '([{}])', this should be considered correct as it includes the proper closing brackets.
\end{minted}
\end{tcolorbox}

\vspace{0.5em}
\noindent\textbf{Mask Generation.}  
For prompt segmentation, we generate masked variants to localize ineffective segments based on their contribution to loss:

\begin{tcolorbox}[title={\textbf{\small Prompt for Mask Generation}}, boxrule=2pt, arc=0mm, breakable]
\begin{minted}[fontsize=\scriptsize, breaklines, frame=lines, framesep=2mm, tabsize=4, style=vs, autogobble]{python}
Given the following prompt, analyze it to identify up to 5 relatively independent units (segments) that are not tightly connected to their surrounding content. Such units can include individual methods, rules, or examples. For each unit, consider whether masking (removing) it would leave the surrounding prompt logically coherent and understandable.

Your task:
1. Carefully read the current prompt: {prompt}
2. If possible, segment the prompt into up to 5 independent units. Only select units whose removal would not disrupt the overall flow or meaning of the prompt.
3. For each selected unit, wrap it in <mask></mask> tags.
4. Wrap the entire prompt in <prompt></prompt> tags.
5. If you find the prompt does not contain any truly independent units suitable for masking, simply output the prompt wrapped in <prompt></prompt> without any <mask></mask> tags.

Formatting requirements:
- Do not exceed 5 masked segments in total.
- Do not create overlapping or nested masks.
- Each masked unit should represent a coherent, removable segment (such as a method, rule, or example), not a random phrase.
- Maintain all other prompt content unchanged.

Example input:
current prompt: {prompt}

Example output with masking:
<prompt> ... Some instructions ... <mask>This is an example rule to be masked.</mask> ... More instructions ... </prompt>

Example output with no masking:
<prompt> ... Full original prompt (no <mask> tags) ... </prompt>
\end{minted}
\end{tcolorbox}

\noindent This prompt is used during the mask-guided analysis phase in PMPO to identify local prompt segments whose removal yields significant loss changes. These segments are considered candidates for targeted rewriting in subsequent iterations.
\subsection{Experiment Details}

\subsubsection{Tasks and Data Details}
\label{appendix:data}

We evaluate PMPO across multiple datasets covering symbolic reasoning, math problem-solving, and instruction-following. Table~\ref{tab:dataset} summarizes dataset sizes and splits used in our experiments.

\begin{table}[htbp]
\caption{Dataset sizes and data splits used for training and evaluation.}
\label{tab:dataset}
\renewcommand\tabcolsep{3.2pt}
\renewcommand\arraystretch{1.2}
\small
\setlength{\abovecaptionskip}{0.1cm}
\setlength{\belowcaptionskip}{-0.2cm}
\centering
\begin{tabular}{l|cc}
\hline
\textbf{Dataset Name} & \textbf{Test Size} & \textbf{Train (max)} \\
\hline
BBH* & 6,511 & 1304 \\
GSM8K & 1,319 & 50 \\
AQUA-RAT & 254 & 50 \\
AlpacaEval 2.0 & 805 & 50 \\
\hline
\end{tabular}
\end{table}

\subparagraph{BBH*}
The BIG-Bench Hard (BBH) benchmark~\cite{suzgun2022challenging} comprises 23 challenging tasks selected from the BIG-Bench suite, focusing on areas where language models previously underperformed compared to average human raters. Each task includes 250 test examples, totaling 6,511 samples. We utilize the full BBH dataset for evaluation and randomly sample 50 examples for training.

\subparagraph{GSM8K}
GSM8K~\cite{cobbe2021training} is a math word problem benchmark requiring multi-step numerical reasoning. We use the standard test split (1,319 samples) for evaluation. For training, we randomly sample up to 50 examples from the training set.

\subparagraph{AQUA-RAT}
AQUA-RAT~\cite{ling2017program} contains multiple-choice math questions requiring algebraic reasoning and textual comprehension. Following prior work, we use the full test set (2,371 questions) and randomly sample 50 training examples for optimization.

\subparagraph{AlpacaEval 2.0}
AlpacaEval 2.0~\cite{dubois2025lengthcontrolledalpacaevalsimpleway} is a benchmark for evaluating instruction-following ability of language models using GPT-4 Turbo as the judge. The dataset contains 805 diverse prompts spanning tasks such as open-ended generation, roleplay, summarization, and reasoning. We use the full evaluation set for testing. For prompt optimization training, we use preference-labeled pairs (selected vs. rejected responses) from the public repository \texttt{reciprocate/alpaca-eval}, which offers high-quality alignment signals derived from automatic LLM feedback. From this set, we sample 50 training pairs with clear preference margins. These examples allow PMPO to optimize prompts for alignment-style objectives using cross-entropy on preferred vs. dispreferred completions.

\begin{flushleft}
\textbf{Note:} * BBH refers to the complete set of 23 tasks in the BIG-Bench Hard benchmark.
\end{flushleft}

\begin{itemize}
    \item \textbf{Boolean Expressions:} Evaluate the truth value of a random Boolean expression consisting of Boolean constants (True, False) and basic Boolean operators (and, or, and not).
    \item \textbf{Causal Judgment:} Given a short story (involving moral, intentional, or counterfactual analysis), determine how a typical person would answer a causal question about the story.
    \item \textbf{Date Understanding:} Given a small set of sentences about a particular date, answer the provided question (e.g., “The concert was scheduled to be on 06/01/1943, but was delayed by one day to today. What is the date yesterday in MM/DD/YYYY?”).
    \item \textbf{Disambiguation QA:} Given a sentence with an “ambiguous” pronoun, either determine whether the sentence is inherently ambiguous or state the antecedent of the pronoun.
    \item \textbf{Dyck Languages:} Predict the sequence of the closing parentheses of a Dyck-4 word without its last few closing parentheses.
    \item \textbf{Formal Fallacies Syllogisms Negation:} Determine whether an argument—presented informally—can be logically deduced from a provided set of statements.
    \item \textbf{Geometric Shapes:} Given a full SVG path element containing multiple commands, determine the geometric shape that would be generated.
    \item \textbf{Hyperbaton (Adjective Ordering):} Given two English-language sentences, determine the one with the correct adjective order.
    \item \textbf{Logical Deduction:} Deduce the order of a sequence of objects based on clues about their spatial relationships and placements.
    \item \textbf{Movie Recommendation:} Given a list of movies a user might have watched and liked, recommend a new, relevant movie from a list of candidates.
    \item \textbf{Multi-Step Arithmetic:} Solve multi-step equations involving basic arithmetic operations.
    \item \textbf{Navigate:} Given a series of navigation steps, determine whether the agent would return to the starting point.
    \item \textbf{Object Counting:} Given a list of items and quantities, determine the count of a certain object class (e.g., fruits).
    \item \textbf{Penguins in a Table:} Given a table of penguins (and sometimes new information), answer a question about the penguins' attributes.
    \item \textbf{Reasoning about Colored Objects:} Answer a simple question about the color of an object based on a given context.
    \item \textbf{Ruin Names:} Given an artist, band, or movie name, identify a one-character edit that changes the meaning humorously.
    \item \textbf{Salient Translation Error Detection:} Given a German sentence and its English translation, determine the type of translation error present.
    \item \textbf{Snarks:} Given two nearly-identical sentences, determine which one is sarcastic.
    \item \textbf{Sports Understanding:} Determine whether a fictitious sports-related sentence is plausible.
    \item \textbf{Temporal Sequences:} Given a person's daily schedule, determine when they could perform another activity.
    \item \textbf{Tracking Shuffled Objects:} Given initial object positions and a series of swaps, determine final positions.
    \item \textbf{Web of Lies:} Evaluate the truth value of a Boolean function expressed as a word problem.
    \item \textbf{Word Sorting:} Given a list of words, sort them lexicographically.
\end{itemize}

\subsubsection{Configuration}
\label{sec:config_baselines}

We compare ours against two categories of baseline methods: manually designed prompting strategies and automated prompt optimization frameworks. The former apply structural heuristics to improve reasoning, while the latter leverage LLMs or search algorithms to generate and refine prompts automatically.

\paragraph{Manually Designed Prompting Strategies}

\subparagraph{Chain-of-Thought (CoT)~\cite{cot}.}
CoT prompting enhances reasoning by encouraging the model to generate intermediate steps before the final answer. It improves performance in arithmetic and multi-step tasks by including phrases like ``Let's think step by step'' to elicit structured reasoning chains.

\subparagraph{Step-Back~\cite{stepback}.}
Step-Back prompting first guides the model to abstract high-level concepts before applying them to the specific task. This abstraction step enables more principled reasoning, particularly in STEM tasks, by helping the model organize relevant knowledge before execution.

\subparagraph{Rephrase-and-Respond~\cite{rephrase}.}
Rephrase-and-Respond (RaR) improves answer quality by prompting the model to reformulate the input question before solving it. This internal clarification reduces ambiguity and enhances robustness, especially in under-specified or complex queries.

\paragraph{Automated Prompt Optimization Approaches}

\subparagraph{OPRO~\cite{opro}.}
OPRO treats prompt design as a black-box optimization problem. An LLM iteratively proposes and evaluates new prompts based on previous ones and their accuracy on training examples. Although effective, it depends on repeated generation and evaluation over full datasets.

\subparagraph{EvoPrompt~\cite{evoprompt}.}
EvoPrompt combines evolutionary algorithms with LLMs to maintain and evolve a population of prompts through mutation and selection. New prompts are generated via LLM-guided variation, and top-performing variants are retained across generations.

\subparagraph{PromptWizard~\cite{PromptWizard}.}
PromptWizard uses a critique-and-synthesis loop where one LLM analyzes prompt weaknesses and another generates refinements. This self-reflective mechanism incrementally improves prompts based on failure analysis, with high sample efficiency and task specialization.

\subparagraph{TextGrad~\cite{textguard}.}
TextGrad performs optimization via pseudo-gradients derived from LLM-generated feedback. It applies gradient descent-like updates to prompts, reverting bad updates using validation feedback, and supports optimization without requiring explicit supervision labels.

\subparagraph{PromptBreeder~\cite{promptbreeder}.}
PromptBreeder evolves both task prompts and mutation strategies in parallel. It uses co-evolution to generate increasingly effective prompts without human intervention, achieving state-of-the-art performance on complex reasoning benchmarks.

\subparagraph{SPO~\cite{SPO}.}
SPO avoids human-labeled data by comparing outputs from multiple prompts and selecting better-performing variants through self-judgment. This self-supervised loop incrementally refines prompts based on model preference rather than external metrics.

\subparagraph{PMPO (Probabilistic Metric Prompt Optimization).}
Ours introduces a unified loss-based prompt optimization framework. It evaluates prompt quality using token-level cross-entropy or preference loss, requiring only forward passes without output sampling. In each iteration, ours identifies low-utility prompt segments via model-adaptive masking, generates variants based on hard examples, and selects improved candidates via likelihood minimization. Unlike black-box or introspection-based methods, ours supports both supervised and preference tasks under a consistent evaluation scheme and demonstrates scalability to smaller models with limited data.

\subsubsection{Baseline Prompt}
\label{appendix:baseline-prompt}

In this section, we present the baseline prompts used for comparison across all methods. For fairness, all prompt optimization approaches that require an initial prompt use the same \textbf{COT Prompt} as the starting point.

\begin{tcolorbox}[title={\textbf{\small AO Prompt}}, boxrule=2pt, arc=0mm, breakable]
\begin{minted}[fontsize=\scriptsize, breaklines, breakanywhere, frame=lines, framesep=2mm, tabsize=4, style=vs, autogobble]{markdown}
Please output directly and only output the final answer.
\end{minted}
\end{tcolorbox}

\begin{tcolorbox}[title={\textbf{\small COT Prompt}}, boxrule=2pt, arc=0mm, breakable]
\begin{minted}[fontsize=\scriptsize, breaklines, breakanywhere, frame=lines, framesep=2mm, tabsize=4, style=vs, autogobble]{markdown}
Let's think step by step.
\end{minted}
\end{tcolorbox}

\begin{tcolorbox}[title={\textbf{\small Step-back Prompt}}, boxrule=2pt, arc=0mm, breakable]
\begin{minted}[fontsize=\scriptsize, breaklines, breakanywhere, frame=lines, framesep=2mm, tabsize=4, style=vs, autogobble]{markdown}
Please first think about the principles involved in solving this task which could be helpful.
And then provide a solution step by step for this question.
\end{minted}
\end{tcolorbox}

\begin{tcolorbox}[title={\textbf{\small RaR Prompt}}, boxrule=2pt, arc=0mm, breakable]
\begin{minted}[fontsize=\scriptsize, breaklines, breakanywhere, frame=lines, framesep=2mm, tabsize=4, style=vs, autogobble]{markdown}
Please rephrase the question in a way that is easier to understand, minimizing ambiguity and considering edge cases.
And then provide a solution step by step for the question.
\end{minted}
\end{tcolorbox}

\subsection{Case Study}

\begin{figure}[htbp]
    \centering
    \includegraphics[width=0.75\linewidth]{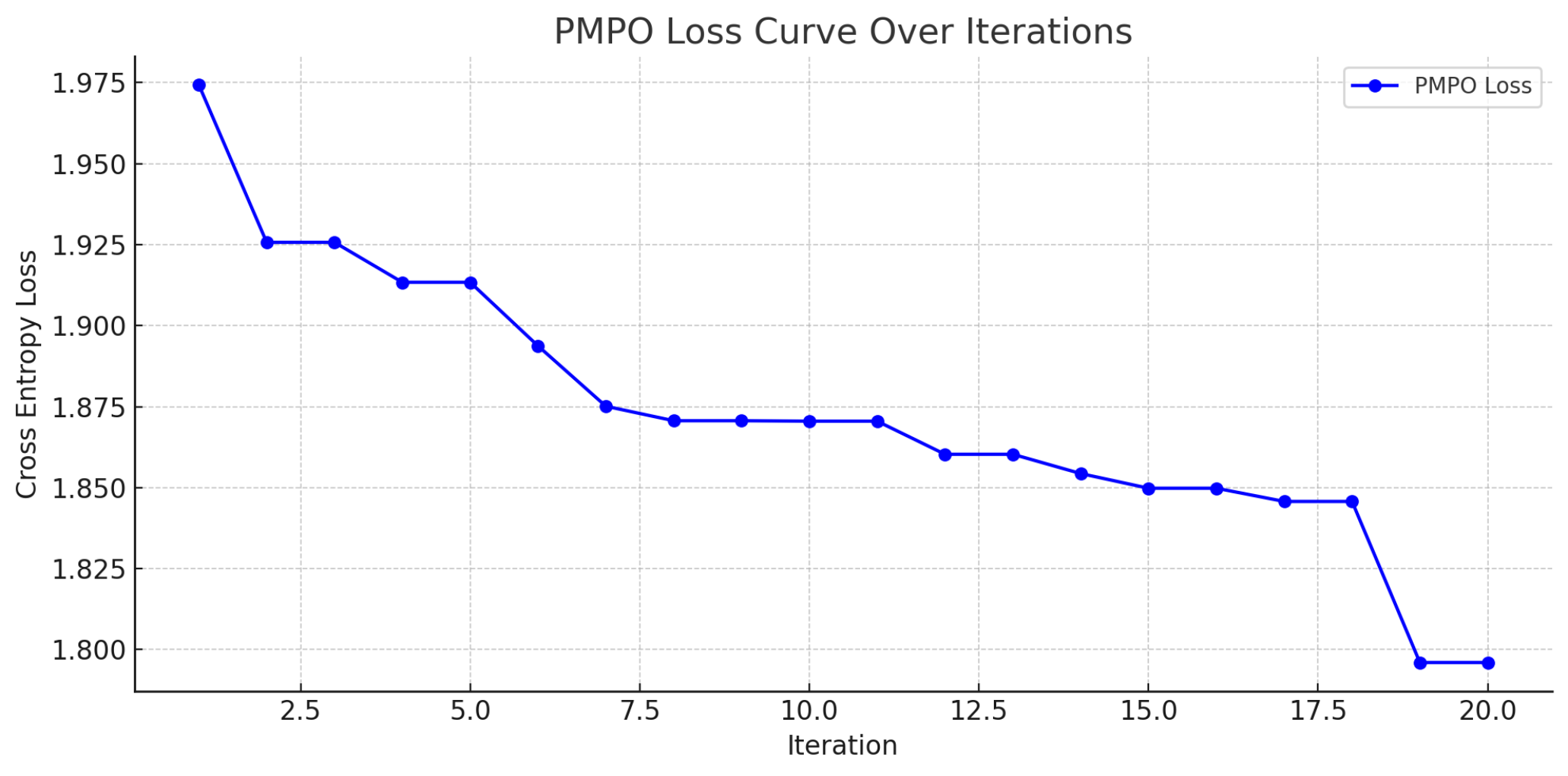}
    \caption{Cross-entropy loss per iteration during prompt optimization on AQUA-RAT.}
    \label{fig:loss-aqua}
\end{figure}

To better understand how PMPO progressively improves prompt quality during optimization, we conduct a case study on the AQUA-RAT dataset. Starting with the base prompt \texttt{Let's think step by step}, we observe how PMPO rewrites the prompt over 20 iterations by identifying high-loss examples and refining the prompt to better capture general mathematical reasoning patterns. Table~\ref{fig:loss-aqua} shows selected steps in this evolution along with the corresponding cross-entropy losses. As observed, the prompts gradually become more structured and specific in mathematical reasoning and format expectations.

Figure~\ref{fig:loss-aqua} shows the loss curve, highlighting the performance gains across the 20 iterations. Notably, early iterations achieve rapid improvements, while later refinements focus on polishing structure and clarity.

\begin{tcolorbox}[title={\textbf{\small Iter 1, Loss = 1.974}}, boxrule=2pt, arc=0mm, breakable]
\begin{minted}[fontsize=\scriptsize, breaklines]{markdown}
Let's think step by step
\end{minted}
\end{tcolorbox}

\begin{tcolorbox}[title={\textbf{\small Iter 2, Loss = 1.926}}, boxrule=2pt, arc=0mm, breakable]
\begin{minted}[fontsize=\scriptsize, breaklines]{markdown}
Given a mathematical problem involving factors and divisibility, your task is to determine which of the provided options correctly identifies all numbers that the given positive whole number must be divisible by. Follow these steps to solve the problem effectively:

1. Identify the prime factors of the given number based on the information provided.
2. Determine the least common multiple (LCM) of these prime factors, which represents the smallest number that the given number must be divisible by.
3. Evaluate each option against the LCM to check if it meets the divisibility criteria.
4. Select the correct option(s) that accurately reflect the divisibility rules derived from the prime factors.

Ensure your solution clearly explains your reasoning process and highlights why certain options are correct or incorrect based on the divisibility properties of the given number.
\end{minted}
\end{tcolorbox}

\begin{tcolorbox}[title={\textbf{\small Iter 4, Loss = 1.913}}, boxrule=2pt, arc=0mm, breakable]
\begin{minted}[fontsize=\scriptsize, breaklines]{markdown}
Given a mathematical problem involving geometric shapes or numerical properties, your task is to solve the problem and identify the correct answer among the provided options. Here are the steps to follow for an effective solution:

1. Carefully read and comprehend the problem statement, identifying the key elements and requirements.
2. Apply relevant mathematical formulas, principles, or theorems to derive a solution. This may involve calculations, logical deductions, or both.
3. Evaluate each provided option in light of your solution to determine which one accurately reflects the correct answer.
4. Clearly explain your reasoning process and highlight why certain options are correct or incorrect based on the problem's specific requirements.

Maintain precision in your calculations and ensure that your solution aligns with the core mathematical concepts involved. Use clear and concise language to articulate your thought process and conclusions.
\end{minted}
\end{tcolorbox}

\begin{tcolorbox}[title={\textbf{\small Iter 6, Loss = 1.894}}, boxrule=2pt, arc=0mm, breakable]
\begin{minted}[fontsize=\scriptsize, breaklines]{markdown}
To effectively solve the given math problem and select the correct answer option, follow these steps:

1. Carefully read and understand the problem statement to identify all relevant conditions and relationships among the numbers involved.
2. Apply mathematical principles and formulas related to the conditions identified to derive necessary calculations or conclusions.
3. Analyze each answer option in light of the derived conclusions and evaluate whether it satisfies the conditions of the problem.
4. Clearly explain the rationale behind selecting the correct answer option, detailing how the option meets the specified conditions.

Ensure your response includes a clear explanation of the reasoning process used to arrive at the correct answer, demonstrating a thorough understanding of the problem's underlying mathematical concepts.
\end{minted}
\end{tcolorbox}

\begin{tcolorbox}[title={\textbf{\small Iter 8, Loss = 1.871}}, boxrule=2pt, arc=0mm, breakable]
\begin{minted}[fontsize=\scriptsize, breaklines]{markdown}
Given a mathematical problem involving numerical properties, your task is to solve the problem and identify the correct answer among the provided options. Follow these steps for an effective solution:

1. Thoroughly read and understand the problem statement, identifying key elements such as given factors, divisibility rules, or other numerical properties.
2. Apply appropriate mathematical principles, theorems, or logical deductions to derive a solution. This may include factorization, divisibility tests, or other relevant calculations.
3. Evaluate each provided option against your derived solution to determine the correct answer.
4. Clearly explain your reasoning process, highlighting why certain options are correct or incorrect based on the problem's specific requirements.

Ensure your solution is precise and aligned with the core mathematical concepts involved. Use clear, concise language to articulate your thought process and conclusions.

By following these guidelines, you will be able to effectively solve the problem and select the correct answer from the given options.
\end{minted}
\end{tcolorbox}

\begin{tcolorbox}[title={\textbf{\small Iter 10, Loss = 1.871}}, boxrule=2pt, arc=0mm, breakable]
\begin{minted}[fontsize=\scriptsize, breaklines]{markdown}
Given a mathematical problem involving geometric or numerical properties, your task is to accurately solve the problem and identify the correct answer from the provided options. To achieve this, follow these steps:

1. Carefully read the problem statement to identify all given elements and required calculations, such as dimensions, formulas, or numerical relationships.
2. Apply relevant mathematical principles, formulas, or logical reasoning to derive a solution. This could involve using geometric formulas, algebraic equations, or numerical analysis.
3. Compare your solution with each provided option to pinpoint the correct answer.
4. Provide a clear explanation of your reasoning, detailing how you arrived at your solution and why certain options are valid or invalid.

Your solution should be precise, demonstrating a thorough understanding of the underlying mathematical concepts. Use clear and concise language to articulate your thought process and justify your conclusions.

By adhering to these guidelines, you will be able to effectively solve the problem and confidently select the correct answer from the given options.
\end{minted}
\end{tcolorbox}

\begin{tcolorbox}[title={\textbf{\small Iter 12, Loss = 1.860}}, boxrule=2pt, arc=0mm, breakable]
\begin{minted}[fontsize=\scriptsize, breaklines]{markdown}
Given a mathematical problem involving numerical properties or relationships, your task is to accurately solve the problem and identify the correct answer from the provided options. Follow these steps to ensure a comprehensive and accurate solution:

1. **Comprehend the Problem**: Read the problem statement carefully to identify all given elements and the required calculations, including any numerical relationships, formulas, or conditions.
2. **Apply Mathematical Principles**: Utilize appropriate mathematical concepts, formulas, or logical reasoning to solve the problem. This may include applying algebraic equations, numerical analysis, or other relevant mathematical tools.
3. **Verify the Solution**: Compare your derived solution with each provided option to determine the correct answer.
4. **Explain Your Reasoning**: Provide a clear and detailed explanation of your thought process and the steps taken to arrive at your solution. Justify why certain options are valid or invalid based on your calculations and reasoning.

Your response should demonstrate a deep understanding of the mathematical concepts involved and use clear, concise language to explain your approach and conclusions.

By following these guidelines, you will be able to effectively solve the problem and confidently choose the correct answer from the given options.
\end{minted}
\end{tcolorbox}

\begin{tcolorbox}[title={\textbf{\small Iter 14, Loss = 1.854}}, boxrule=2pt, arc=0mm, breakable]
\begin{minted}[fontsize=\scriptsize, breaklines]{markdown}
Given a mathematical problem involving divisibility rules or factor relationships, follow these steps to find the correct answer:

1. **Identify the Core Elements**: Clearly define the key numbers and conditions provided in the problem statement.
2. **Apply Mathematical Principles**: Use divisibility rules and factorization techniques to deduce the required conditions.
3. **Evaluate Each Option**: Carefully examine each given option against the derived conditions to determine which ones satisfy the problem's criteria.
4. **Select the Correct Answer**: Based on your evaluation, choose the most accurate set of options that fulfill the problem's requirements.

Ensure your solution is logically sound and aligns with fundamental mathematical concepts. Avoid making assumptions not supported by the given information. 

Your response should include a clear explanation of your reasoning process and the final selected answer.
\end{minted}
\end{tcolorbox}

\begin{tcolorbox}[title={\textbf{\small Iter 15, Loss = 1.850}}, boxrule=2pt, arc=0mm, breakable]
\begin{minted}[fontsize=\scriptsize, breaklines]{markdown}
Given a mathematical problem, such as finding the curved surface area of a geometric shape, follow these steps to accurately solve the problem and select the correct answer from the provided options:

1. **Understand the Problem**: Carefully read and comprehend the given problem, identifying all relevant numerical values and geometric properties mentioned.
2. **Recall Relevant Formulas**: Identify and recall the appropriate mathematical formulas or principles necessary to solve the problem.
3. **Calculate the Solution**: Apply the identified formulas to the given data, performing all necessary calculations step-by-step.
4. **Verify Each Option**: Compare your calculated result with each provided answer option to determine which one matches your solution.
5. **Choose the Correct Answer**: Select the option that correctly represents the solution to the problem based on your calculations.

Ensure your solution process is logically consistent and grounded in fundamental mathematical principles. Avoid making unsupported assumptions and verify each step of your calculation.

Your response should include a detailed rationale explaining your reasoning process, leading to the selection of the correct answer.
\end{minted}
\end{tcolorbox}

\begin{tcolorbox}[title={\textbf{\small Iter 17, Loss = 1.846}}, boxrule=2pt, arc=0mm, breakable]
\begin{minted}[fontsize=\scriptsize, breaklines]{markdown}
Given a mathematical problem, such as calculating the curved surface area of a geometric shape, follow these steps to accurately solve the problem and select the correct answer from the provided options:

1. **Analyze the Problem Statement**: Carefully read the problem statement to identify all relevant numerical values and geometric properties involved.
2. **Identify Applicable Formulas**: Recall and list the relevant mathematical formulas or principles that are pertinent to solving the problem.
3. **Perform Calculations**: Apply the identified formulas to the given data, ensuring each step of the calculation is accurate and logically sound.
4. **Evaluate Answer Options**: Compare your calculated result against each provided answer option to determine the correct match.
5. **Select the Correct Answer**: Choose the answer option that precisely corresponds to your calculated solution.

In your response, provide a detailed rationale that outlines your reasoning process, clearly demonstrating how you arrived at your solution. Ensure that your approach adheres to fundamental mathematical principles and avoids any unsupported assumptions. Verify each step of your calculation for accuracy and consistency.
\end{minted}
\end{tcolorbox}

\begin{tcolorbox}[title={\textbf{\small Iter 19, Loss = 1.796}}, boxrule=2pt, arc=0mm, breakable]
\begin{minted}[fontsize=\scriptsize, breaklines]{markdown}
Given a mathematical problem involving geometric shapes, such as finding the curved surface area, follow these guidelines to accurately solve the problem and select the correct answer from the provided options:

1. **Understand the Problem**: Thoroughly comprehend the given problem statement to identify all pertinent numerical values and geometric characteristics.
2. **Recall Relevant Formulas**: Identify and list the applicable mathematical formulas or principles needed to solve the problem.
3. **Execute Calculations**: Apply the recalled formulas to the given data, ensuring each calculation step is precise and logically consistent.
4. **Compare with Options**: Match your calculated result against each provided answer option to pinpoint the correct choice.
5. **Confirm Selection**: Select the answer option that aligns exactly with your calculated solution.

Provide a detailed rationale explaining your reasoning process, highlighting how you applied fundamental mathematical principles to arrive at your solution. Ensure your approach is free from unsupported assumptions and verify each calculation step for accuracy and consistency.
\end{minted}
\end{tcolorbox}

Even with the use of mask-based analysis and model-in-the-loop rewriting for supervision, PMPO does not guarantee improvement in every single iteration. In cases where no generated candidate outperforms the current prompt, the original prompt is retained as the starting point for the next round of optimization.

The degree of improvement across steps may vary due to factors like sampling randomness, batch composition, and the inherent difficulty of refining an already strong prompt. Unlike model fine-tuning, prompt generation lacks a stable optimization trajectory, making step-to-step changes inherently more variable.

\begin{table}[h!]
\centering
\caption{Performance comparison on the BBH benchmark using the smaller Qwen2.5-7B model. PMPO (Ours) achieves the highest average accuracy and outperforms other methods on the majority of tasks. Bold values indicate the best-performing method(s) for each task.}
\label{tab:qwen7b_results}
\resizebox{\textwidth}{!}{%
\begin{tabular}{@{}lcccccccc@{}}
\toprule
\textbf{Task} & \textbf{AO} & \textbf{PromptWizard} & \textbf{StepBack} & \textbf{CoT} & \textbf{OPRO} & \textbf{EvoPrompt} & \textbf{RaR} & \textbf{Ours} \\ \midrule
boolean\_expressions & 0.840 & 0.904 & 0.936 & 0.892 & 0.868 & 0.804 & \textbf{0.964} & 0.940 \\
causal\_judgement & 0.572 & 0.503 & 0.599 & 0.487 & 0.503 & 0.567 & 0.583 & \textbf{0.615} \\
date\_understanding & 0.544 & 0.536 & 0.616 & 0.580 & 0.636 & 0.620 & 0.604 & \textbf{0.652} \\
disambiguation\_qa & 0.592 & 0.640 & \textbf{0.700} & 0.648 & 0.660 & 0.664 & 0.616 & 0.644 \\
dyck\_languages & \textbf{0.392} & 0.248 & 0.168 & 0.148 & 0.180 & 0.176 & 0.156 & 0.320 \\
formal\_fallacies & 0.596 & 0.612 & 0.728 & 0.724 & \textbf{0.748} & \textbf{0.748} & 0.660 & 0.740 \\
geometric\_shapes & 0.304 & 0.440 & 0.624 & 0.596 & 0.612 & 0.616 & 0.608 & \textbf{0.652} \\
hyperbaton & 0.568 & 0.632 & 0.808 & 0.808 & 0.784 & 0.824 & 0.772 & \textbf{0.828} \\
movie\_recommendation & \textbf{0.604} & 0.584 & 0.444 & 0.352 & 0.388 & 0.412 & 0.368 & 0.592 \\
multistep\_arithmetic\_two & 0.044 & 0.972 & 0.928 & 0.976 & 0.984 & 0.948 & 0.944 & \textbf{0.988} \\
navigate & 0.680 & 0.700 & 0.872 & 0.872 & 0.848 & 0.912 & 0.900 & \textbf{0.920} \\
object\_counting & 0.484 & 0.520 & 0.800 & \textbf{0.824} & 0.636 & 0.804 & 0.672 & 0.816 \\
penguins\_in\_a\_table & 0.541 & 0.568 & 0.904 & 0.856 & 0.897 & \textbf{0.904} & 0.863 & \textbf{0.904} \\
reasoning\_about\_colored\_objects & 0.588 & \textbf{0.948} & 0.940 & 0.896 & 0.908 & 0.924 & 0.908 & \textbf{0.948} \\
ruin\_names & 0.596 & 0.608 & 0.592 & 0.596 & 0.636 & 0.492 & 0.552 & \textbf{0.784} \\
salient\_translation & 0.444 & 0.516 & 0.496 & 0.504 & 0.512 & \textbf{0.648} & 0.520 & 0.620 \\
snarks & 0.747 & 0.640 & \textbf{0.860} & 0.753 & 0.584 & 0.747 & 0.680 & 0.742 \\
sports\_understanding & 0.648 & 0.084 & 0.536 & 0.324 & \textbf{0.720} & 0.296 & 0.248 & 0.584 \\
temporal\_sequences & 0.652 & 0.736 & 0.688 & 0.608 & 0.736 & 0.656 & 0.604 & \textbf{0.760} \\
tracking\_shuffled\_objects & 0.188 & 0.269 & 0.809 & \textbf{0.853} & 0.845 & 0.832 & 0.764 & 0.829 \\
logical\_deduction & 0.589 & 0.641 & 0.756 & 0.747 & 0.753 & \textbf{0.765} & 0.739 & 0.761 \\
web\_of\_lies & 0.576 & 0.500 & 0.844 & \textbf{0.916} & 0.676 & 0.912 & 0.384 & 0.836 \\
word\_sorting & 0.348 & 0.464 & 0.324 & 0.228 & 0.448 & 0.272 & 0.488 & \textbf{0.552} \\ \midrule
\textbf{Average Accuracy} & 0.531 & 0.582 & 0.705 & 0.687 & 0.704 & 0.698 & 0.648 & \textbf{0.733} \\
\textbf{Best performing tasks} & 2 & 1 & 3 & 3 & 1 & 3 & 1 & \textbf{13} \\ \bottomrule
\end{tabular}%
}
\end{table}
\subsection{Further Analysis}
\label{sec:appendix_further_analysis}

\subsubsection{Qualitative Analysis of Sample Efficiency}

Our claim of \textbf{high sample efficiency} is rooted in PMPO’s computational design and the granularity of its evaluation signal. Computationally, unlike methods such as PromptWizard or OPRO that require generating full outputs via costly autoregressive decoding, PMPO evaluates prompts by computing token-level likelihoods through a single forward pass for each sample. This bypasses the expensive decoding step, making the optimization loop significantly faster and more scalable. Furthermore, the optimization signal in PMPO is the cross-entropy loss, which captures per-token deviations from the target output rather than relying on a binary correctness score. This fine-grained signal provides a more informative gradient, allowing the framework to learn more effectively from fewer samples.

However, the number of samples remains a critical factor. When optimizing on only one sample, PMPO is prone to overfitting to lexical artifacts or task-specific details. For instance, in a sentiment classification task, training on a single negative review might yield an overly rigid instruction like:
\begin{quote}
“If the reviewer says the filmmakers tried hard but the movie seems awfully sloppy, and mentions many factual errors about Lucille Ball, respond with ‘Negative’. Otherwise, respond with ‘Positive’.”
\end{quote}
Such a prompt fails to generalize. In contrast, when optimizing over a small batch of top-k examples ($k \ge 3$), we observe that the learned prompts consistently capture generalized patterns aligned with the underlying task (e.g., abstract reasoning steps or classification criteria). This suggests PMPO can achieve stable optimization with very few samples (e.g., 3–5) by leveraging the cross-entropy signal across a small but diverse set of examples to avoid overfitting.

\subsubsection{Performance Evaluation on Smaller Models}

To further demonstrate the empirical reliability and scalability of PMPO across different model sizes, we conducted a comprehensive evaluation on a smaller model, Qwen2.5-7B. Table~\ref{tab:qwen7b_results} shows that PMPO continues to outperform strong baselines across the diverse tasks in the BBH benchmark, achieving the highest average accuracy.

Notably, the instruction rewrites generated by methods like PromptWizard tend to be significantly more verbose. This verbosity can lead to performance degradation in smaller models, which may struggle with capacity constraints and overly specified instructions. These findings further emphasize PMPO’s generalizability and robustness, particularly in resource-constrained settings where concise and effective prompts are crucial.

\subsubsection{Complementary Effects with Few-Shot Prompting}
\label{sec:appendix_few_shot}

PMPO and n-shot prompting are not mutually exclusive; rather, they target complementary dimensions of prompt design. Few-shot prompting conveys task format and solution patterns through concrete examples, while PMPO focuses on refining the core instruction to improve its clarity, coverage, and alignment with the task's intent. The benefits of both approaches can be additive, with PMPO enhancing the instruction’s generalizability and few-shot examples providing task-specific grounding.

To empirically validate this relationship, we conducted experiments on the BBH benchmark under various few-shot configurations, using both a standard base prompt and a PMPO-optimized prompt. The results are summarized in Table~\ref{tab:pmpo_vs_nshot}.

\begin{table}[h!]
\centering
\caption{Comparison of a base prompt and a PMPO-optimized prompt across different numbers of few-shot examples ($n$). Scores are averaged over BBH tasks. The results show that for every value of $n$, the PMPO-optimized prompt outperforms the base prompt.}
\label{tab:pmpo_vs_nshot}
\begin{tabular}{@{}ccc@{}}
\toprule
\textbf{Number of Shots ($n$)} & \textbf{Prompt Type} & \textbf{Average Score} \\
\midrule
\multirow{2}{*}{0}  & Base             & 73.75                  \\
                    & PMPO  & \textbf{79.21 }                 \\ \cmidrule(l){1-3}
\multirow{2}{*}{1}  & Base             & 75.02                  \\
                    & PMPO  & \textbf{80.69 }                 \\ \cmidrule(l){1-3}
\multirow{2}{*}{3}  & Base             & 76.69                  \\
                    & PMPO  & \textbf{81.11 }                 \\ \cmidrule(l){1-3}
\multirow{2}{*}{5}  & Base             & 76.89                  \\
                    & PMPO  & \textbf{82.19}                  \\ \cmidrule(l){1-3}
\multirow{2}{*}{7}  & Base             & 76.58                  \\
                    & PMPO  & \textbf{81.59}                  \\ \cmidrule(l){1-3}
\multirow{2}{*}{9}  & Base             & 77.97                  \\
                    & PMPO  & \textbf{82.37}         \\ \cmidrule(l){1-3}
\multirow{2}{*}{16} & Base             & 78.17                  \\
                    & PMPO  & \textbf{82.23}                  \\ \bottomrule
\end{tabular}
\end{table}

The results lead to several key observations. First, the PMPO-optimized prompt in a zero-shot setting (79.21) already outperforms the base prompt with 16-shots (78.17), demonstrating the substantial impact of instruction optimization alone. Second, combining PMPO with few-shot examples consistently yields the highest scores, confirming the additive nature of the two techniques. Finally, performance does not monotonically increase with the number of shots; for example, the 7-shot performance is lower than the 5-shot performance in both settings. This suggests that an excessive number of examples can introduce noise or cause instruction drift. In contrast, PMPO systematically improves the base prompt's structure, making the instruction more interpretable and robust to input variation. Thus, PMPO offers a significant improvement over few-shot prompting alone, and the two approaches are best utilized in tandem.

\subsection{Prompt Optimized by ours}
\label{appendix:optimized-prompts}

In this section, we present the optimized prompts produced by our method during the main experiments. All prompts were optimized using Qwen2.5-14B as the optimization model, with the original model also serving as the evaluation and execution model. For brevity, we show representative examples for each benchmark dataset.

\paragraph{BBH (Big-Bench Hard)}
For PMPO training on BBH, we treat the correct answer option as the \textit{selected} output, and the remaining distractor options as \textit{rejected}, enabling preference-based supervision using the model's likelihoods.
\begin{tcolorbox}[title={\textbf{\small BBH Prompt (boolean\_expressions)}}, boxrule=2pt, arc=0mm, breakable]
\begin{minted}[fontsize=\scriptsize, breaklines, breakanywhere, frame=lines, framesep=2mm, tabsize=4, style=vs, autogobble]{markdown}
Your task is to accurately evaluate the truth value of a given Boolean expression. This expression may include Boolean constants (True, False) and fundamental Boolean operators (and, or, and not). Adhere strictly to standard Boolean logic throughout this process.

### Key Considerations:
- **Parentheses Handling**: Nested conditions within parentheses require careful management. Start by evaluating the innermost parentheses and proceed outward.
- **Operator Precedence**: Observe the correct order of operations: "not" has the highest precedence, followed by "and", and then "or". This ensures proper simplification of the expression.
- **Logical Reasoning**: Clearly articulate the step-by-step logical reasoning applied during the evaluation process. Maintain clarity without sacrificing depth.

### Input Format:
A string containing a Boolean expression, such as "not not True and True and not True".
\end{minted}
\end{tcolorbox}

\begin{tcolorbox}[title={\textbf{\small BBH Prompt (causal\_judgement)}}, boxrule=2pt, arc=0mm, breakable]
\begin{minted}[fontsize=\scriptsize, breaklines, breakanywhere, frame=lines, framesep=2mm, tabsize=4, style=vs, autogobble]{markdown}
Given a short story involving moral, intentional, or counterfactual analysis, your task is to determine how a typical person would respond to a causal question about the story. Your response should provide a clear chain of thought that leads to a reasoned conclusion.

Begin by thoroughly reading the story and the causal question. Identify the key elements and the specific causal relationship highlighted in the question. Analyze the cause-and-effect dynamics from a typical person's perspective. Formulate your answer by clearly stating whether a typical person would consider the proposed causal link valid, supported by a logical rationale.

Guidelines for responding:
1. Carefully read and understand the story and the causal question.
2. Identify the critical causal factors within the story.
3. Evaluate the causal relationship from the viewpoint of a typical person.
4. Provide a logical and well-supported explanation for your answer.
5. Ensure your response comprehensively addresses the task without being overly specific to any particular example.
\end{minted}
\end{tcolorbox}

\begin{tcolorbox}[title={\textbf{\small BBH Prompt (date\_understanding)}}, boxrule=2pt, arc=0mm, breakable]
\begin{minted}[fontsize=\scriptsize, breaklines, breakanywhere, frame=lines, framesep=2mm, tabsize=4, style=vs, autogobble]{markdown}
To effectively solve date-related questions based on given narratives, follow this structured yet flexible approach:

### Objective:
Determine the correct date from a brief narrative containing specific details about dates and events. Provide the final answer in the "MM/DD/YYYY" format, accompanied by a detailed chain-of-thought explanation.

### Detailed Instructions:

#### Step 1: Comprehend the Narrative
Read the provided sentences carefully to identify all relevant dates and events. Pay special attention to any changes in dates due to scheduling adjustments or other factors.

#### Step 2: Pinpoint Key Dates
Identify all explicit dates mentioned in the narrative. Also, consider any shifts in these dates caused by external conditions.

#### Step 3: Evaluate Required Calculations
Determine whether the question necessitates adding or subtracting days, weeks, months, or years from the key dates. Remember to factor in calendar nuances, including the varying lengths of months and the impact of leap years.

#### Step 4: Execute Calculations
Perform the required calculations accurately, taking into account complexities such as leap years and differences in month lengths. Ensure your calculations are precise and logical.

#### Step 5: Deliver the Result
Present the final answer in the "MM/DD/YYYY" format. Provide a clear rationale explaining each step of your reasoning process and any assumptions made.
\end{minted}
\end{tcolorbox}

\begin{tcolorbox}[title={\textbf{\small BBH Prompt (disambiguation\_qa)}}, boxrule=2pt, arc=0mm, breakable]
\begin{minted}[fontsize=\scriptsize, breaklines, breakanywhere, frame=lines, framesep=2mm, tabsize=4, style=vs, autogobble]{markdown}
Given a sentence that contains a potentially ambiguous pronoun, your goal is to determine whether the pronoun clearly refers to a specific noun within the sentence or if its meaning is uncertain due to insufficient context. This process involves identifying the most likely referent of the pronoun or recognizing its ambiguity, providing a thorough rationale for your conclusion.

To accomplish this task effectively, follow these steps:

1. **Identify the Pronoun**: Locate the pronoun in the sentence.
2. **Analyze Context**: Carefully examine the sentence for any clues that might reveal the pronoun's antecedent. Consider the logical relationship between the pronoun and all potential antecedents.
3. **Decide on Clarity or Ambiguity**: Based on your analysis, determine if the pronoun's reference is clear or ambiguous.
4. **Justify Your Conclusion**: Provide a clear and well-supported explanation for your decision, referencing pertinent details from the sentence.

Your response should conform to one of the following formats:

- For sentences deemed ambiguous: "The sentence is ambiguous because [detailed reasoning based on context]."
- For sentences where the pronoun's reference is clear: "The pronoun refers to [specific noun] because [detailed rationale based on context]."
\end{minted}
\end{tcolorbox}

\begin{tcolorbox}[title={\textbf{\small BBH Prompt (dyck\_languages)}}, boxrule=2pt, arc=0mm, breakable]
\begin{minted}[fontsize=\scriptsize, breaklines, breakanywhere, frame=lines, framesep=2mm, tabsize=4, style=vs, autogobble]{markdown}
Your task is to predict and complete the sequence of closing parentheses to form a valid Dyck-4 word. A Dyck-4 word requires a balanced arrangement of four types of parentheses: round (), square [], curly {}, and angle <>. Each opening bracket must be correctly matched with its corresponding closing bracket, ensuring proper nesting and balance throughout the sequence.

Follow these guidelines to solve the problem effectively:

1. **Identify Unmatched Brackets**: Start by identifying all opening brackets in the input sequence that do not yet have corresponding closing brackets. This step is essential for determining the types of closing brackets required next.
2. **Prioritize Innermost Structures**: Always give priority to closing the innermost unmatched brackets first, respecting the hierarchical matching rules of the different types of brackets.
3. **Ensure Balance**: Maintain the balance between the counts of each type of opening and closing brackets. This ensures that each closing bracket is placed appropriately and maintains the overall balance of the sequence.
4. **Continuous Validation**: After adding each closing bracket, continuously validate that the sequence remains balanced and properly nested.
5. **Explain Chain-of-Thought**: Include a detailed Chain-of-Thought section that explains your reasoning at each step, clearly demonstrating how you ensured compliance with the Dyck-4 word criteria.

Your response should include:
- The completed sequence of closing parentheses.
- A detailed Chain-of-Thought explanation outlining the decision-making process.
\end{minted}
\end{tcolorbox}

\begin{tcolorbox}[title={\textbf{\small BBH Prompt (formal\_fallacies)}}, boxrule=2pt, arc=0mm, breakable]
\begin{minted}[fontsize=\scriptsize, breaklines, breakanywhere, frame=lines, framesep=2mm, tabsize=4, style=vs, autogobble]{markdown}
Given a context that includes a series of logical statements and premises, your task is to evaluate whether a presented argument can be logically deduced from the given context. Your response should clearly indicate whether the argument is valid or invalid, supported by a detailed explanation of your reasoning process.

To approach this task effectively, follow these steps:

1. Carefully analyze each premise to understand the relationships and conditions established.
2. Compare the argument's conclusion with the premises. Assess whether the conclusion logically follows from the premises provided.
3. If the argument is valid, outline how each premise contributes to the logical progression leading to the conclusion. If invalid, specify which premise(s) or logical step(s) fail to support the conclusion adequately.

Guidelines for your response:
- Clearly state your answer as either "valid" or "invalid".
- Provide a concise but thorough explanation of your reasoning process.
- Use clear and precise language to articulate your thoughts.
\end{minted}
\end{tcolorbox}

\begin{tcolorbox}[title={\textbf{\small BBH Prompt (geometric\_shapes)}}, boxrule=2pt, arc=0mm, breakable]
\begin{minted}[fontsize=\scriptsize, breaklines, breakanywhere, frame=lines, framesep=2mm, tabsize=4, style=vs, autogobble]{markdown}
Your task is to interpret an SVG path element specified within a "d" attribute and deduce the geometric shape that would be formed upon executing this path. This process involves understanding a variety of path commands such as "M" for moving to a point, "L" for drawing lines, and other commands according to the SVG standard.

To achieve this, follow these steps:

1. **Command Analysis**: Carefully analyze the "d" attribute to identify various commands and their corresponding parameters. Recognize how each command contributes to forming the overall path.

2. **Path Sequence Understanding**: Grasp the sequence and relationships between commands and points. Note how movements and connections between points influence the final shape's structure.

3. **Geometric Shape Identification**: Based on the parsed commands and parameters, deduce the geometric shape represented by the path. Consider both simple and complex shapes that could emerge from the given path commands.

4. **Shape Matching**: Choose the most suitable geometric shape from the provided options. Ensure your selection accurately reflects the interpreted path data without simplifying or misinterpreting it.

Remember, the path may describe intricate shapes beyond basic polygons. Your analysis should accurately capture the complexity of the path data to determine the precise geometric form.
\end{minted}
\end{tcolorbox}

\begin{tcolorbox}[title={\textbf{\small BBH Prompt (hyperbaton)}}, boxrule=2pt, arc=0mm, breakable]
\begin{minted}[fontsize=\scriptsize, breaklines, breakanywhere, frame=lines, framesep=2mm, tabsize=4, style=vs, autogobble]{markdown}
Your task is to evaluate two given English sentences and determine which one correctly follows the standard adjective order in English. To accomplish this, adhere to the following guidelines:

1. **Adjective Order Principle**: In English, adjectives typically follow a specific order: Opinion, Size, Age, Shape, Color, Origin, Material, Purpose.

2. **Analyze Adjectives**: Carefully scrutinize each sentence to identify and note the adjectives and their positions relative to the nouns they modify. Pay special attention to their sequence.

3. **Compare Sequences**: Utilize the established order to assess and compare the sequences of adjectives in both sentences. Highlight any discrepancies from the correct order.

4. **Select the Correct Sentence**: Choose the sentence that accurately adheres to the English adjective order rules.

**Response Format**: Indicate your chosen sentence using a single letter (A or B).
\end{minted}
\end{tcolorbox}

\begin{tcolorbox}[title={\textbf{\small BBH Prompt (logical\_deduction\_five\_objects)}}, boxrule=2pt, arc=0mm, breakable]
\begin{minted}[fontsize=\scriptsize, breaklines, breakanywhere, frame=lines, framesep=2mm, tabsize=4, style=vs, autogobble]{markdown}
Given a description of several objects arranged in a specific order based on their spatial relationships, determine the correct sequence of these objects. 

**Task Description**: Your primary goal is to deduce the precise order of a series of objects based on the clues provided about their relative positions. The clues may involve terms like "left," "right," "above," "below," or any other directional indicators relevant to the arrangement of the objects. Ensure that the solution aligns perfectly with all given clues without contradiction.

**Guidelines**:
1. Carefully read through the entire description to fully understand the spatial relationships between the objects.
2. Use a systematic approach to map out the positions of each object according to the clues.
3. Consider multiple perspectives if necessary to verify the accuracy of your deduction.
4. If the sequence is ambiguous after applying the clues, re-examine the descriptions for additional hints or rephrase the clues for clarity.
5. Provide a clear and concise answer detailing the order of the objects from left to right or top to bottom, depending on the orientation described.

**Example Format**: Your answer should clearly state the order of the objects as determined by the clues.
\end{minted}
\end{tcolorbox}

\begin{tcolorbox}[title={\textbf{\small BBH Prompt (logical\_deduction\_seven\_objects)}}, boxrule=2pt, arc=0mm, breakable]
\begin{minted}[fontsize=\scriptsize, breaklines, breakanywhere, frame=lines, framesep=2mm, tabsize=4, style=vs, autogobble]{markdown}
Your task is to accurately determine the sequential order of a series of objects based on the provided clues about their spatial relationships and placements. This requires meticulous analysis and logical deduction to arrange the objects in their specified sequence.

### Guidelines for Solving:
1. **Comprehend Clues**: Thoroughly read and understand each statement to grasp the positional relationships among the objects.
2. **Direct Position Identification**: Identify clues that specify the exact position of an object (e.g., "Object X is the leftmost").
3. **Relative Position Interpretation**: Analyze clues indicating the relative positions of objects (e.g., "Object Y is three spots before Object Z").
4. **Elimination Process**: Use the process of elimination to determine the correct positions for objects once some placements are known.
5. **Solution Verification**: Ensure that your final sequence aligns with all given clues without any conflicts.

### Problem-Solving Strategy:
- Start by identifying any direct placement clues that definitively place objects.
- Apply relative positioning clues to establish relationships between objects.
- Utilize elimination techniques to fill in the remaining positions.
- Confirm your sequence to make sure it satisfies all provided clues.

### Logical Reasoning Framework:
Develop a systematic approach to analyze and resolve the clues:
1. **Initial Setup**: List all objects and note any direct placements from the clues.
2. **Relationship Mapping**: Outline the relationships between objects as described by the clues.
3. **Reasoning Deduction**: Combine the mapped relationships with elimination techniques to deduce the order.
4. **Validation Check**: Review your sequence to ensure it adheres to all clue requirements.

### Response Structure:
Provide your answer as an ordered list of objects, accompanied by a detailed explanation of your reasoning process.
\end{minted}
\end{tcolorbox}

\begin{tcolorbox}[title={\textbf{\small BBH Prompt (logical\_deduction\_three\_objects)}}, boxrule=2pt, arc=0mm, breakable]
\begin{minted}[fontsize=\scriptsize, breaklines, breakanywhere, frame=lines, framesep=2mm, tabsize=4, style=vs, autogobble]{markdown}
Your task is to deduce the precise order of a series of objects based on the provided clues about their spatial relationships and placements. Each scenario will detail a set of objects along with statements indicating their relative positions or rankings. Your goal is to accurately establish the exact order of these objects according to the orientation described in the input.

**Instructions:**
1. Thoroughly examine the input paragraph that describes the objects and their relationships.
2. Utilize the given clues to logically deduce the correct sequence of the objects.
3. Present the objects in their correct order, starting with the object at the highest rank or position as per the clues.
4. If the input lacks sufficient information to determine a unique order, clearly state that the order cannot be definitively established based on the provided clues.

**Guidelines:**
- Refrain from making assumptions that go beyond what is explicitly stated in the input.
- Ensure your response is concise and focuses exclusively on identifying and listing the correct order of objects, without additional commentary or explanations unless required by the task.
- Scenarios may involve various types of objects and different numbers of objects, but the core principle remains consistent: use the provided clues to determine the correct order.
\end{minted}
\end{tcolorbox}

\begin{tcolorbox}[title={\textbf{\small BBH Prompt (movie\_recommendation)}}, boxrule=2pt, arc=0mm, breakable]
\begin{minted}[fontsize=\scriptsize, breaklines, breakanywhere, frame=lines, framesep=2mm, tabsize=4, style=vs, autogobble]{markdown}
Given a list of movies a user has enjoyed, your task is to recommend a new, relevant movie from a set of four potential choices. Follow the steps below to ensure a thoughtful and accurate recommendation:

1. **Understand User Preferences:** Carefully examine the given list of movies to identify common themes, genres, and styles that the user appreciates. This analysis will form the basis of the user's taste profile.

2. **Evaluate Movie Choices:** Compare each of the four movie options against the user's taste profile. Pay attention to genre, theme, director, actors, and overall style. Aim to find a movie that closely aligns with the user's preferences, avoiding any significant deviations.

3. **Choose the Optimal Match:** Select the movie that best matches the user's preferences, demonstrating a deep understanding of both the user's tastes and the distinct characteristics of each movie option.

**Instructions for Response:**
- Provide only the letter corresponding to the recommended movie from the four options.

**Guiding Principles:**
- Highlight thematic, genre, and stylistic consistency with the user's preferences.
- Avoid suggesting movies that diverge markedly from the user's established tastes.
- Exhibit a refined understanding of the user's preferences and the unique qualities of each movie option.

**Goal:**
Increase user satisfaction by recommending a movie that closely reflects their established preferences.
\end{minted}
\end{tcolorbox}

\begin{tcolorbox}[title={\textbf{\small BBH Prompt (multistep\_arithmetic\_two)}}, boxrule=2pt, arc=0mm, breakable]
\begin{minted}[fontsize=\scriptsize, breaklines, breakanywhere, frame=lines, framesep=2mm, tabsize=4, style=vs, autogobble]{markdown}
Your task is to accurately solve a complex multi-step arithmetic problem that involves a variety of operations, such as addition, subtraction, multiplication, and potentially division. It's important to pay close attention to parentheses, as they determine the order of operations according to standard mathematical principles.

To tackle this problem effectively and precisely, follow these steps:

1. Start by evaluating all expressions within parentheses, beginning with the deepest ones. This ensures that you resolve the operations inside the parentheses before proceeding to other parts of the equation.
2. After resolving all parentheses, move on to multiplication and division operations from left to right. This step adheres to the standard order of operations.
3. Lastly, perform addition and subtraction operations from left to right until you reach the final result. Carefully execute each operation to prevent errors.
4. Clearly present your final answer, supported by a detailed, step-by-step explanation of your calculations to ensure transparency and accuracy.

When crafting your response, make sure to:
- Break down the problem systematically, addressing each component sequentially and methodically.
- Provide a precise and professionally articulated explanation, with each calculation step clearly documented and logically explained.
\end{minted}
\end{tcolorbox}

\begin{tcolorbox}[title={\textbf{\small BBH Prompt (navigate)}}, boxrule=2pt, arc=0mm, breakable]
\begin{minted}[fontsize=\scriptsize, breaklines, breakanywhere, frame=lines, framesep=2mm, tabsize=4, style=vs, autogobble]{markdown}
Given a set of navigation instructions for an agent, your task is to determine if the agent ends up back at its initial starting point. Follow these structured guidelines to methodically analyze the navigation instructions:

**Guidelines for Analyzing Navigation Instructions:**
1. **Initial Setup**: Position the agent at the Origin (Point 0) and orient it Facing Forward.
2. **Command Handling**:
   - **Movement Commands**: Adjust the agent's position based on its orientation (Forward, Backward).
   - **Orientation Commands**: Alter the agent's direction (Turn around, Turn left, Turn right) without changing its position.
3. **Sequential Execution**:
   - Process each command one by one.
   - Update the agent's position and orientation after executing each command.
4. **Final Position Verification**:
   - After all commands have been executed, check if the agent's position matches the initial Origin.
   - Determine if the agent has returned to the starting point.
5. **Response Format**:
   - Give chain of thought.
   - Provide a clear "Yes" or "No" answer.
   - Include a detailed explanation of your reasoning process.
\end{minted}
\end{tcolorbox}

\begin{tcolorbox}[title={\textbf{\small BBH Prompt (object\_counting)}}, boxrule=2pt, arc=0mm, breakable]
\begin{minted}[fontsize=\scriptsize, breaklines, breakanywhere, frame=lines, framesep=2mm, tabsize=4, style=vs, autogobble]{markdown}
Your task is to calculate the total quantity of items belonging to a specific category from a provided list of possessions and their quantities. This involves identifying the target category, categorizing each item, and summing up the quantities of those that fit the criteria.

Follow these steps to complete the task:

1. **Understand the Target Category**: Carefully read the question to understand which category of items needs to be counted (e.g., fruits, vegetables, musical instruments).

2. **Review the List of Possessions**: Go through the list of items and their quantities provided in the input.

3. **Categorize Each Item**: For every item listed, decide whether it falls under the specified category.

4. **Sum Up Quantities**: Add together the quantities of all items that belong to the target category.

5. **Present the Result**: Clearly state the total count of items in the specified category.
\end{minted}
\end{tcolorbox}

\begin{tcolorbox}[title={\textbf{\small BBH Prompt (penguins\_in\_a\_table)}}, boxrule=2pt, arc=0mm, breakable]
\begin{minted}[fontsize=\scriptsize, breaklines, breakanywhere, frame=lines, framesep=2mm, tabsize=4, style=vs, autogobble]{markdown}
Given a unique table containing detailed information about penguins (and sometimes additional data), your task is to accurately answer questions about the attributes of the penguins listed in the table. Ensure that your response strictly relies on the provided data without incorporating any external information.

Each table includes essential columns such as "name," "age," "height (cm)," and "weight (kg)," potentially supplemented with other relevant details. Each row corresponds to a distinct penguin, highlighting their individual characteristics.

To approach this task effectively, follow these steps:
1. Thoroughly examine the provided table(s) to understand the structure and content.
2. Clearly identify the question posed about the penguins' attributes.
3. Utilize the data from the table to deduce the correct answer.
4. If additional information is provided, integrate it seamlessly into your analysis.
5. Deliver your answer precisely and succinctly, ensuring it accurately addresses the question.
\end{minted}
\end{tcolorbox}

\begin{tcolorbox}[title={\textbf{\small BBH Prompt (reasoning\_about\_colored\_objects)}}, boxrule=2pt, arc=0mm, breakable]
\begin{minted}[fontsize=\scriptsize, breaklines, breakanywhere, frame=lines, framesep=2mm, tabsize=4, style=vs, autogobble]{markdown}
Given a detailed description of objects placed on a surface, your goal is to accurately identify and answer a question about the color of a specific object within that description. The response should be concise and directly answer the question, optionally including a brief rationale if requested.

To accomplish this task efficiently and effectively, follow these steps:

1. **Understand the Description**: Carefully read through the provided description to familiarize yourself with each object and its corresponding color.
2. **Pinpoint the Object of Interest**: Identify which object's color is being queried in the question.
3. **Retrieve the Color Information**: Refer back to your initial notes to locate the color associated with the object of interest.
4. **Provide a Clear Response**: Directly state the color of the object in a straightforward manner.
5. **Offer Optional Explanation**: If asked, briefly explain your reasoning process.
\end{minted}
\end{tcolorbox}

\begin{tcolorbox}[title={\textbf{\small BBH Prompt (ruin\_names)}}, boxrule=2pt, arc=0mm, breakable]
\begin{minted}[fontsize=\scriptsize, breaklines, breakanywhere, frame=lines, framesep=2mm, tabsize=4, style=vs, autogobble]{markdown}
Given an artist, band, or movie name, your goal is to craft a single-character edit that transforms the meaning into something humorous. Follow these guidelines to ensure your response is both creative and effective:

1. **Input Understanding**: Carefully read the provided name, whether it's an artist, band, or movie title.
2. **Creative Transformation**: Modify the name by changing exactly one character. The change should significantly alter the meaning in a comedic manner.
3. **Humor Explanation**: For each edit, provide a short explanation highlighting why the alteration is funny. Consider creating a play on words, introducing an absurd situation, or crafting a pun.
4. **Avoid Repetition**: Ensure that each edit is unique and adds a fresh perspective to the name.

**Execution Strategy**:
- **Character Impact Analysis**: Evaluate how changing each character affects the overall meaning and potential for humor.
- **Phonetic & Literal Changes**: Consider both sound-based and literal alterations that could create a humorous effect.
- **Cultural References**: Think about popular culture, idioms, and common phrases that could be humorously altered with a single character change.

**Guidance for Humorous Edits**:
- Strive for humor that appeals to a wide audience while also offering depth for those who enjoy sophisticated wordplay.
- Ensure the edits are clear and understandable, avoiding ambiguity that might detract from the humor.
\end{minted}
\end{tcolorbox}

\begin{tcolorbox}[title={\textbf{\small BBH Prompt (salient\_translation\_error\_detection)}}, boxrule=2pt, arc=0mm, breakable]
\begin{minted}[fontsize=\scriptsize, breaklines, breakanywhere, frame=lines, framesep=2mm, tabsize=4, style=vs, autogobble]{markdown}
Given a German sentence and its English translation, your task is to carefully analyze the translation for errors and categorize them into one of the following six types:

- **Named Entities**: Any change to names, places, or other identifiable entities.
- **Numerical Values**: Errors involving incorrect changes to numbers, dates, or units.
- **Modifiers or Adjectives**: Improper modifications to descriptive terms associated with nouns.
- **Negation or Antonyms**: Issues related to the introduction or removal of negations, or the incorrect transformation of comparatives into their opposites.
- **Facts**: Trivial factual inaccuracies that do not fit into the aforementioned categories.
- **Dropped Content**: Omission of a key component of the sentence in the translation.

To complete this task effectively, follow these structured steps:

1. Carefully read and comprehend both the German source sentence and its English translation.
2. Perform a detailed comparison between the two sentences to identify any discrepancies.
3. Categorize these discrepancies according to the predefined error types.
4. Select the most appropriate category for the identified error(s).
\end{minted}
\end{tcolorbox}

\begin{tcolorbox}[title={\textbf{\small BBH Prompt (snarks)}}, boxrule=2pt, arc=0mm, breakable]
\begin{minted}[fontsize=\scriptsize, breaklines, breakanywhere, frame=lines, framesep=2mm, tabsize=4, style=vs, autogobble]{markdown}
To effectively distinguish between two nearly identical sentences and identify the one that uses sarcasm, follow the enhanced guidelines below:

### Objective
Your goal is to recognize which of the two given sentences is intended to convey sarcasm. Sarcasm is characterized by language that expresses a meaning opposite to its literal sense, often used to mock or criticize.

### Analytical Approach
1. **Analyze Word Choice and Sentence Structure**: Carefully examine the words and structure of each sentence for subtle cues that hint at sarcasm.
2. **Imagery of Real-Life Context**: Visualize a plausible scenario in which the sentences might be said to better understand the speaker's intent.
3. **Tone Evaluation**: Assess whether the tone of each sentence suggests sincerity or mockery/criticism.
4. **Examine Literal vs. Implied Meaning**: Look for significant discrepancies between the literal interpretation and the implied message, which often signals sarcasm.
5. **Identify Signs of Irony**: Pay attention to any exaggerated elements or unexpected twists that indicate irony or mockery.

### Response Format
Provide a clear indication of which sentence is sarcastic and support your conclusion with a rationale focusing on ironic or mocking elements.
\end{minted}
\end{tcolorbox}

\begin{tcolorbox}[title={\textbf{\small BBH Prompt (sports\_understanding)}}, boxrule=2pt, arc=0mm, breakable]
\begin{minted}[fontsize=\scriptsize, breaklines, breakanywhere, frame=lines, framesep=2mm, tabsize=4, style=vs, autogobble]{markdown}
Given a fictitious sentence related to sports, evaluate its plausibility. Your evaluation should follow these steps:

1. **Comprehend the Sentence**: Thoroughly read and understand the sentence, taking note of the sport or activity it might relate to, the key actors involved, and any specific details about the action or conditions described.

2. **Analyze Key Components**: Identify the central elements of the sentence, including the athlete's name, the action performed, and any particular specifications like body part used or environmental factors that may influence the action.

3. **Assess Logical Possibility**: Determine if the described scenario fits within the logical and practical boundaries of sports. Consider the typical rules, physical constraints, and standard practices relevant to the sport implied by the sentence.

4. **Justify Your Conclusion**: Clearly explain your judgment regarding the plausibility of the sentence. Use specific references from the sentence to support your reasoning, detailing how these elements align with or deviate from established norms in sports.

5. **Detail the Reasoning Process**: Provide a detailed Chain-of-Thought that outlines each step of your reasoning process. This should include initial observations, intermediate deductions, and the final justification for your conclusion.
\end{minted}
\end{tcolorbox}

\begin{tcolorbox}[title={\textbf{\small BBH Prompt (temporal\_sequences)}}, boxrule=2pt, arc=0mm, breakable]
\begin{minted}[fontsize=\scriptsize, breaklines, breakanywhere, frame=lines, framesep=2mm, tabsize=4, style=vs, autogobble]{markdown}
Given a detailed account of a person's daily schedule, including various activities and their corresponding times, your objective is to identify potential periods of availability throughout the day when the person could have engaged in an additional activity. Follow these steps to achieve this:

1. **Comprehend the Schedule**: Carefully review the provided sequence of events and note down the start and end times for each activity. Ensure you capture all pertinent details about the day's beginning and end, along with any specific operational hours or restrictions related to the locations involved.

2. **Detect Intervals of Availability**: Examine the timeline for intervals where no activity is specified. These gaps suggest potential windows of availability for engaging in additional activities.

3. **Apply Contextual Constraints**: Take into account additional contextual factors such as daily routines, operational hours of places (like parks or shops), and other known limitations or rules (such as curfews).

4. **Define Availability Periods**: Clearly outline the identified periods of availability in your response, making sure they align with the provided schedule and any supplementary context.

5. **Use Concise Language**: Respond using clear, professional language focused exclusively on the periods of availability without adding extraneous details.

6. **Systematic Examination**: Employ a thorough, step-by-step approach to ensure all potential gaps in the schedule are analyzed accurately.
\end{minted}
\end{tcolorbox}

\begin{tcolorbox}[title={\textbf{\small BBH Prompt (tracking\_shuffled\_objects\_five\_objects)}}, boxrule=2pt, arc=0mm, breakable]
\begin{minted}[fontsize=\scriptsize, breaklines, breakanywhere, frame=lines, framesep=2mm, tabsize=4, style=vs, autogobble]{markdown}
To effectively determine the final positions of a set of entities after a series of pairwise swaps, adopt a clear, structured, and flexible approach focusing on precision, accuracy, and comprehensive coverage. This method aims to streamline the process of tracking transformations while ensuring thorough adherence to the task's requirements.

### Enhanced Problem-Solving Framework

#### Step 1: Define Initial States
1. Identify and label each entity uniquely (e.g., using names, labels).
2. Document the initial position or state of each entity clearly to establish a baseline for tracking changes.

#### Step 2: Execute Swaps Systematically
1. For each swap in the sequence:
   - Identify the two entities involved.
   - Update their positions or states according to the swap.
   - Record the swap chronologically to ensure traceability and validation.

#### Step 3: Confirm Final States
1. After applying all swaps:
   - Verify the final positions or states of all entities.
   - Cross-reference the final configuration against the recorded swaps to ensure accuracy and completeness.

### Core Principles
- **Sequential Order**: Apply swaps in the order provided to maintain the integrity of the transformation sequence.
- **Clear Documentation**: Use consistent methods for documenting position or state updates to reduce errors and enhance traceability.
- **Thorough Validation**: Conduct rigorous checks on the final positions or states to ensure they accurately reflect the intended outcomes based on the swaps executed.

### Practical Application Examples
This framework can be applied to various situations, such as:
- Students exchanging books.
- Dancers swapping partners.
- Entities undergoing pairwise exchanges in any context.

The fundamental requirement remains constant: transitioning from an initial setup to a final configuration through a series of swaps.

#### Detailed Task Execution
1. **Define Initial Setup**:
   - Clearly outline the starting positions or states of all entities.
2. **Implement Sequential Swaps**:
   - Apply each swap systematically, updating positions or states as specified.
3. **Validate Final Configuration**:
   - Thoroughly verify the final positions or states to ensure they correctly reflect the sequence of swaps executed.
\end{minted}
\end{tcolorbox}

\begin{tcolorbox}[title={\textbf{\small BBH Prompt (tracking\_shuffled\_objects\_seven\_objects)}}, boxrule=2pt, arc=0mm, breakable]
\begin{minted}[fontsize=\scriptsize, breaklines, breakanywhere, frame=lines, framesep=2mm, tabsize=4, style=vs, autogobble]{markdown}
Your task involves determining the final positions of a set of entities after undergoing a series of pairwise swaps. To effectively tackle this task, adhere to the following guidelines designed to ensure accuracy and efficiency:

1. **Initial Position Identification**: Clearly define and list each entity alongside its initial position. Employ a structured format such as a table or list to facilitate easy identification and ensure a clear understanding of the starting configuration.

2. **Detailed Recording of Swaps**: Closely document each swap operation, specifying which entities are involved and the order in which these swaps take place. This meticulous documentation is essential for accurate tracking and proper application of each swap.

3. **Step-by-Step Swap Execution**: Execute each swap operation strictly in the recorded sequence. Update the positions of the entities after every swap, maintaining an organized and clear record to ensure precision and adherence to the swap order.

4. **Final Position Compilation**: Upon completing all swap operations, compile and present the final positions of all entities in a clear and accessible format.

To enhance your approach, consider the following strategies:
- Utilize a structured format like a table or list to keep detailed records of positions at each stage, thereby increasing clarity and minimizing errors.
- Stress the significance of meticulous documentation and sequential execution of swap operations to maintain accuracy and consistency.
- Implement a flexible yet robust method capable of efficiently managing diverse swap sequences and varying quantities of entities.

Key points to remember:
- Maintain precision and thoroughness when dealing with multiple pieces of information concurrently.
- Ensure your methodology is adaptable to various scenarios involving differing numbers of entities and swap sequences.
- Avoid dependence on specific examples; instead, cultivate a versatile approach applicable across different contexts.
\end{minted}
\end{tcolorbox}

\begin{tcolorbox}[title={\textbf{\small BBH Prompt (tracking\_shuffled\_objects\_three\_objects)}}, boxrule=2pt, arc=0mm, breakable]
\begin{minted}[fontsize=\scriptsize, breaklines, breakanywhere, frame=lines, framesep=2mm, tabsize=4, style=vs, autogobble]{markdown}
Given a set of objects and a sequence of pairwise swaps, the goal is to determine the final positions of each object after all swaps have been executed. The solution should be generalizable, capable of handling any number of objects and any sequence of swaps, without being constrained to specific examples.

### Objective
Determine the final position of each object following a series of pairwise swaps.

### Input Format
- The initial positions of objects (e.g., "Object A starts at position X").
- A list of swaps (e.g., "Swap Object A with Object B").

### Output Format
- The final positions of all objects after executing all swaps.

### Guidelines for Solving
1. **Initialization**: Begin by defining the starting positions of all objects.
2. **Execution of Swaps**: Apply each swap in the sequence systematically.
   - Identify the pair of objects involved in the swap.
   - Temporarily store the positions of these objects.
   - Perform the swap by exchanging their positions.
3. **Verification**: Confirm the final positions of the objects after processing all swaps.
4. **Presentation**: Clearly state the final positions of all objects.

### Chain-of-Thought Guidance
To solve this problem efficiently and accurately:
1. **Initialization Step**: List out the initial positions of all objects.
2. **Swap Processing Step**: For every swap in the sequence:
   - Note the objects involved in the swap.
   - Store their current positions temporarily.
   - Exchange their positions.
3. **Verification Step**: After completing all swaps, review the final positions of the objects to ensure they align with the sequence of swaps.
4. **Conclusion Step**: Provide a clear statement of the final positions for all objects.
\end{minted}
\end{tcolorbox}

\begin{tcolorbox}[title={\textbf{\small BBH Prompt (web\_of\_lies)}}, boxrule=2pt, arc=0mm, breakable]
\begin{minted}[fontsize=\scriptsize, breaklines, breakanywhere, frame=lines, framesep=2mm, tabsize=4, style=vs, autogobble]{markdown}
To accurately evaluate the truth value of a Boolean function presented as a natural-language word problem, adhere to the following detailed and systematic approach:

1. **Understanding the Core Principle**: Recognize that every individual in the scenario is consistently either truthful or deceitful. A truthful individual will never contradict themselves or anyone known to be truthful, whereas a deceitful individual will always contradict statements of truthfulness.

2. **Identifying Assertions**: Closely examine each individual's claim regarding the honesty or deceitfulness of others. These claims serve as the foundational elements for your logical deduction process.

3. **Progressive Deduction**: Begin with the first confirmed truth or lie and progressively deduce the honesty status of each individual. For each new statement, assess its alignment with the established truths or falsehoods from previous statements.

4. **Maintaining a Chain of Thought Record**: Throughout the analysis, keep a detailed log of your logical progression. Note any contradictions or confirmations encountered during the examination of each statement.

5. **Concluding Evaluation**: After compiling all deductions, determine the truth value of the final statement in question. Ensure that your conclusion aligns with all preceding deductions.

**Key Guidelines for Logical Reasoning**:
- A truthful person's statement accurately reflects reality.
- A liar's statement misrepresents reality.
- Consistency across all statements is essential for establishing the honesty of each individual.

**Step-by-Step Detailed Process**:
- Start with the initial statement and evaluate its truth value.
- For each subsequent statement, carefully examine how it aligns with the established truths or falsehoods from earlier statements.
- Maintain a thorough record of any logical inconsistencies or confirmations throughout the analysis.
- Ensure that your final evaluation is consistent with the entire sequence of deductions made.

**Expected Output Format**:
Provide a comprehensive explanation of your chain of thought leading to your conclusion. Each step should be logically coherent and clearly articulated.
\end{minted}
\end{tcolorbox}

\begin{tcolorbox}[title={\textbf{\small BBH Prompt (word\_sorting)}}, boxrule=2pt, arc=0mm, breakable]
\begin{minted}[fontsize=\scriptsize, breaklines, breakanywhere, frame=lines, framesep=2mm, tabsize=4, style=vs, autogobble]{markdown}
Your task is to organize a list of words into alphabetical order. To accomplish this, provide a detailed chain-of-thought explanation that covers the following aspects: word comparison, handling duplicates, ensuring accuracy, and managing edge cases like empty or excessively long lists. Your response should include the sorted list and a structured thought process, as outlined below:

**Input**: Specify the list of words to be sorted.

**Thought Process**:
- **Initial Setup**: Outline any preliminary actions or configurations needed before sorting begins.
- **Comparison Strategy**: Elaborate on how individual words are compared to determine their order in the list.
- **Dealing with Duplicates**: Explain your approach to manage and handle repeated words within the list.
- **Final Check**: Describe the steps taken to verify the correctness of the sorted list.
- **Edge Cases Consideration**: Address special scenarios such as empty lists or lists containing a very high number of entries.

**Sorted Output**: Display the final list arranged alphabetically.
\end{minted}
\end{tcolorbox}

\paragraph{AQUA-RAT}
AQUA-RAT~\cite{ling2017program} contains multiple-choice math questions requiring algebraic reasoning and textual comprehension. The dataset includes detailed rationales for each question, making it suitable for supervision via explanation. In our experiments, we adopt a non-preference-based optimization setup by treating the step-by-step rationale as the supervision target. This enables PMPO to optimize prompts using cross-entropy loss over full reasoning sequences. 

Additionally, since AQUA-RAT is a multiple-choice task, we also support a preference-based training variant, where the correct answer choice is treated as the \textit{selected} output and the incorrect options as \textit{rejected}, allowing PMPO to operate under both supervised and preference paradigms.

\begin{tcolorbox}[title={\textbf{\small AQUA-RAT Prompt}}, boxrule=2pt, arc=0mm, breakable]
\begin{minted}[fontsize=\scriptsize, breaklines, breakanywhere, frame=lines, framesep=2mm, tabsize=4, style=vs, autogobble]{markdown}
Given a math problem that requires determining a specific value through methods such as algebraic manipulation, geometric reasoning, or other applicable techniques, solve the problem systematically, providing clear explanations for each step to illustrate your reasoning process. Once you have found the solution, select the correct answer from the given options and justify your choice logically. Ensure your approach demonstrates flexibility in addressing various problem types, including quantitative relationships, algebraic expressions, and geometric calculations. Focus on applying fundamental mathematical principles in a broad context without relying on specific examples. Your response should be clear, concise, and professionally framed, guiding the reader through the problem-solving process effectively.
\end{minted}
\end{tcolorbox}

\paragraph{GSM8K}GSM8K~\cite{cobbe2021training} is a benchmark of grade-school math word problems that require multi-step reasoning and precise arithmetic. Since each instance in the dataset comes with a detailed rationale, we treat the full explanation as the supervision target in PMPO training. The optimization objective is to minimize cross-entropy loss over the complete reasoning sequence.

As a result, the optimized prompts tend to preserve the distinctive answer format seen in GSM8K (e.g., \texttt{<<calculation=answer>>}) to better align with the ground truth and reduce token-level loss. This shows that PMPO can adapt prompt style and structure based on dataset-specific signal during training.

\begin{tcolorbox}[title={\textbf{\small GSM8K Prompt}}, boxrule=2pt, arc=0mm, breakable]
\begin{minted}[fontsize=\scriptsize, breaklines, breakanywhere, frame=lines, framesep=2mm, tabsize=4, style=vs, autogobble]{markdown}
To solve any math word problem precisely, follow this structured approach, focusing on clear understanding, logical steps, and accurate calculations:

1. Understand the Problem & Identify the Goal  
   - Briefly restate what the problem is asking (e.g., total amount, difference, remainder, cost).  
   - Determine the main operation(s) involved (e.g., addition, subtraction, multiplication, division, unit conversion).

2. Extract Key Information  
   - List all given numbers, units, and relevant conditions from the problem.

3. Plan the Solution  
   - Define what you need to find.  
   - Outline a clear step-by-step approach to solve it.

4. Step-by-Step Reasoning and Calculation  
   - For each step:  
     1. Explain what you are calculating and why.  
     2. Perform the calculation, using the format `<<calculation=answer>>`, including units if needed.  
     3. If not the final step, explain how this leads to the next one.

5. Conclude and Verify  
   - Confirm that the result answers the original question.  
   - Check consistency in logic, units, and values.

6. Final Answer  
   - Provide only the final result, starting with `####`, and include units if applicable.
\end{minted}
\end{tcolorbox}

\paragraph{AlpacaEval 2.0}For AlpacaEval 2.0~\cite{dubois2025lengthcontrolledalpacaevalsimpleway}, we use the \texttt{reciprocate/alpaca-eval} dataset as our training source. This dataset contains a large number of paired outputs annotated as \textit{selected} (preferred) and \textit{rejected} (less preferred) responses across various instruction-following tasks.

We construct preference-based supervision by treating the \textit{selected} outputs as positive targets and the \textit{rejected} ones as negatives. PMPO is then trained to minimize loss on preferred examples while discouraging prompts that increase likelihood on rejected ones. This setup enables effective instruction tuning without requiring explicit gold answers.

\begin{tcolorbox}[title={\textbf{\small AlpacaEval Prompt}}, boxrule=2pt, arc=0mm, breakable]
\begin{minted}[fontsize=\scriptsize, breaklines, breakanywhere, frame=lines, framesep=2mm, tabsize=4, style=vs, autogobble]{markdown}
To generate an informative, helpful, and accurate response to a user's query, follow these steps carefully:

1. **Comprehend the Query**: Read the user's question or request thoroughly to grasp its essence and identify any key terms or concepts that require attention.
2. **Research Thoroughly**: Utilize your extensive knowledge and research abilities to gather precise and current information that aligns with the user's query.
3. **Structure the Response**: Organize the content logically, prioritizing clarity and conciseness to ensure that the most relevant information is easily accessible to the user.
4. **Enhance with Context**: Provide additional context that can enrich the user's understanding and appreciation of the topic, making the response more valuable and engaging.
5. **Maintain Accuracy**: Double-check all factual information and data included in the response to uphold the highest standards of accuracy and reliability.
6. **Communicate Clearly**: Write in a straightforward and professional tone, using terminology that is appropriate and understandable given the context of the query.
7. **Recommend Further Resources**: Where applicable, suggest additional sources or materials that the user can consult for deeper insights or further exploration of the topic.

**Guiding Principles**:
- Always aim to fully address the user's request, providing a comprehensive and well-rounded response.
- Prioritize user-centric thinking, focusing on the relevance and utility of the information provided.
- Employ a systematic approach to problem-solving, breaking down complex queries into manageable parts and addressing them methodically.
- Encourage engagement and curiosity by offering insightful and thought-provoking content.
\end{minted}
\end{tcolorbox}

\subsubsection{Cross-Model Prompt Analysis}

We compare the final prompts optimized by our PMPO framework across different language models, all targeting the same BBH-Navigate task. The results demonstrate that PMPO consistently tailors its optimizations to the reasoning abilities and strengths of the underlying model, while maintaining adherence to the general task formulation. Below, we present the optimized prompts with no additional examples, as used in our final evaluations.

\begin{tcolorbox}[title={\textbf{\small Qwen2.5-0.5B}}, boxrule=2pt, arc=0mm, breakable]
\begin{minted}[fontsize=\scriptsize, breaklines]{markdown}
To further enhance navigation accuracy, consider implementing additional strategies such as:

1. **Path Reconstruction**: Implement algorithms to reconstruct the user's actual route from start to finish, taking into account potential obstacles, turns, and other navigational factors.
2. **Feedback Loop**: Incorporate real-time feedback loops to continuously adjust the user's course based on their current location and movement status.
3. **Error Correction**: Introduce mechanisms to automatically correct errors in the user's movements, ensuring smooth transitions between different areas of the map.
4. **Dynamic Path Generation**: Develop dynamic paths that adapt to changing conditions, such as adjusting routes based on weather conditions or sudden changes in terrain.
5. **Customizable Paths**: Allow users to specify custom paths or waypoints, which can help them avoid dead ends or reach previously unvisited locations more efficiently.
6. **Real-Time Updates**: Implement real-time updates to the navigation system, allowing users to see how their journey is progressing and make informed decisions about their next moves.

These enhancements not only improve navigation accuracy but also provide a more intuitive and personalized experience for users.
\end{minted}
\end{tcolorbox}

\textit{Analysis:} For smaller models like Qwen2.5-0.5B, PMPO generates prompts that emphasize procedural redundancy and compensatory strategies (e.g., feedback loops, real-time updates). This suggests the model benefits from prompts that scaffold task understanding through expanded heuristics rather than strict symbolic reasoning.

\begin{tcolorbox}[title={\textbf{\small LLaMA3.1-8B}}, boxrule=2pt, arc=0mm, breakable]
\begin{minted}[fontsize=\scriptsize, breaklines]{markdown}
Given a sequence of instructions, determine whether a user returns to the origin after following the instructions. The instructions can include taking steps in a specific direction and turning in a specific direction.

- Use explicit directions (e.g., "north", "south")
- Keep all step sizes uniform (1 unit)
- Define the origin clearly as (0, 0)
- Use structured input like "Take X steps [direction]" or "Turn Y degrees [left/right]"

The user starts at (0, 0) facing north. Analyze whether they return to origin after completing all steps.
\end{minted}
\end{tcolorbox}

\textit{Analysis:} On LLaMA3.1-8B, the optimized prompt prioritizes precision and input standardization. Compared to Qwen2.5-0.5B, this prompt uses less redundancy and more explicit spatial constraints, aligning well with the model's stronger logical parsing ability.

\begin{tcolorbox}[title={\textbf{\small Qwen2.5-14B}}, boxrule=2pt, arc=0mm, breakable]
\begin{minted}[fontsize=\scriptsize, breaklines]{markdown}
Given a set of navigation instructions for an agent, your task is to determine if the agent ends up back at its initial starting point. Follow these structured guidelines to methodically analyze the navigation instructions:

**Guidelines for Analyzing Navigation Instructions:**
1. **Initial Setup**: Position the agent at the Origin (Point 0) and orient it Facing Forward.
2. **Command Handling**:
   - **Movement Commands**: Adjust the agent's position based on its orientation (Forward, Backward).
   - **Orientation Commands**: Alter the agent's direction (Turn around, Turn left, Turn right) without changing its position.
3. **Sequential Execution**:
   - Process each command one by one.
   - Update the agent's position and orientation after executing each command.
4. **Final Position Verification**:
   - After all commands have been executed, check if the agent's position matches the initial Origin.
   - Determine if the agent has returned to the starting point.
5. **Response Format**:
   - Give chain of thought.
   - Provide a clear "Yes" or "No" answer.
   - Include a detailed explanation of your reasoning process.

**Systematic Reasoning Approach**:
- Start with the agent at the Origin and Facing Forward.
- For every movement command, move the agent according to its current orientation.
- For every orientation command, change the agent's direction without modifying its position.
- At the end, assess if the final position corresponds to the Origin.
- Explain your thought process thoroughly to justify the conclusion.
\end{minted}
\end{tcolorbox}

\textit{Analysis:} Qwen2.5-14B benefits from richer instruction structures. PMPO's optimized prompt includes explicit reasoning breakdowns and distinguishes between movement and orientation changes, showcasing the model's ability to follow hierarchical and procedural plans.

\begin{tcolorbox}[title={\textbf{\small Qwen2.5-32B}}, boxrule=2pt, arc=0mm, breakable]
\begin{minted}[fontsize=\scriptsize, breaklines]{markdown}
To effectively determine if a user returns to their origin after following a set of directional instructions, follow this structured approach:
1. **Comprehend the Directions**: Thoroughly review the sequence of movements. Each command includes a direction (forward, backward, left, right) and a step count.
2. **Set Up the Coordinate Plane**: Define a coordinate plane with the origin at (0,0). This plane will help track the user's precise location throughout the journey.
3. **Update Coordinates According to Movements**:
   - Increase the X-coordinate for forward moves by the step count.
   - Decrease the X-coordinate for backward moves by the step count.
   - Decrease the Y-coordinate for leftward moves by the step count.
   - Increase the Y-coordinate for rightward moves by the step count.
4. **Assess the End Position**: After processing all commands, verify the end coordinates. A return to the origin is confirmed when the user's final location is (0,0).
5. **Conclude with Clarity**: Based on the final coordinates, clearly communicate whether the user has reached the starting point again.

This systematic procedure guarantees a precise and clear determination of the user's final position relative to the initial point, addressing the navigation challenge comprehensively.
\end{minted}
\end{tcolorbox}

\textit{Analysis:} The prompt tailored for Qwen2.5-32B reflects its strong symbolic reasoning capacity. PMPO optimizes for Cartesian reasoning by introducing coordinate systems, directional vectors, and arithmetic evaluation—enhancing spatial comprehension without overcomplicating the structure.

\begin{tcolorbox}[title={\textbf{\small DeepSeek-R1-Distill-Qwen-1.5B}}, boxrule=2pt, arc=0mm, breakable]
\begin{minted}[fontsize=\scriptsize, breaklines]{markdown}
To check if someone returns to their origin point, add up all movements: count right steps as positive numbers and left steps as negative numbers. If the sum equals zero, they've returned to the starting point. If not, they've ended elsewhere.

Example: three steps right (+3) and three steps left (-3) gives 0, meaning they returned to origin. Four steps right (+4) and three steps left (-3) gives +1, meaning they didn't.
\end{minted}
\end{tcolorbox}

\textit{Analysis:} For distilled or small-scale models like DeepSeek-R1, PMPO simplifies the logic into linear additive operations. This aligns with the model's limited abstraction depth, ensuring the task is achievable with minimal symbolic overhead.

\end{document}